\documentclass[pdflatex,sn-mathphys-num,iicol]{sn-jnl}

\usepackage{caption}
\usepackage{graphicx}%
\usepackage{multirow}%
\usepackage{amsmath,amssymb,amsfonts}%
\usepackage{amsthm}%
\usepackage{mathrsfs}%
\usepackage[title]{appendix}%
\usepackage{xcolor}%
\usepackage{textcomp}%
\usepackage{manyfoot}%
\usepackage{booktabs}%
\usepackage{algorithm}%
\usepackage{algorithmicx}%
\usepackage{algpseudocode}%
\usepackage{listings}%
\usepackage{multicol}
\usepackage{subcaption}
\usepackage{orcidlink}

\theoremstyle{thmstyleone}%
\theoremstyle{thmstyletwo}%
\theoremstyle{thmstylethree}%

\begin{document}

\title[Article Title]{RobotIQ: Empowering Mobile Robots with Human-Level Planning for Real-World Execution}

\author*[1,2]{\fnm{Emmanuel K.} \sur{Raptis} \orcidlink{0000-0003-0033-7925}}\email{eraptis@ee.duth.gr}

\author[1,2]{\fnm{Athanasios Ch.} \sur{Kapoutsis} \orcidlink{0000-0002-1688-036X}}\email{athakapo@iti.gr}

\author[1,2]{\fnm{Elias B.} \sur{Kosmatopoulos}}\email{kosmatop@ee.duth.gr}

\affil*[1]{\orgdiv{Department of Electrical and Computer Engineering}, \orgname{Democritus University of Thrace}, \orgaddress{\street{Kimmeria Campus}, \city{Xanthi}, \postcode{67100}, \country{Greece}}}

\affil[2]{\orgdiv{Information Technologies Institute}, \orgname{The Centre for Research \& Technology Hellas}, \orgaddress{\street{6th km Harilaou - Thermis}, \city{Thessaloniki}, \postcode{57001}, \country{Greece}}}


\abstract{ This paper introduces RobotIQ, a framework that empowers mobile robots with human-level planning capabilities, enabling seamless communication via natural language instructions through any Large Language Model. The proposed framework is designed in the ROS architecture and aims to bridge the gap between humans and robots, enabling robots to comprehend and execute user-expressed text or voice commands. Our research encompasses a wide spectrum of robotic tasks, ranging from fundamental logical, mathematical, and learning reasoning 
for transferring knowledge in domains
like navigation, manipulation, and object localization, enabling the application of learned behaviors from simulated environments to real-world operations.
All encapsulated within a modular crafted robot library suite of API-wise control functions, RobotIQ offers a fully functional AI-ROS-based toolset that allows researchers to design and develop their own robotic actions tailored to specific applications and robot configurations. The effectiveness of the proposed system was tested and validated both in simulated and real-world experiments focusing on a home service scenario that included an assistive application designed for elderly people. RobotIQ with an open-source, easy-to-use, and adaptable robotic library suite for any robot can be found at \url{https://github.com/emmarapt/RobotIQ}.}



\keywords{Robot Intelligence, Task Planning,  LLM code generation, Robotic Operating System, Symbolic AI}



\maketitle

\section{Introduction}
\label{sec:intro}
Intelligence Quotient (IQ) in robotic platforms typically refers to the cognitive abilities of a robotic system and its level of sophistication in terms of problem-solving, learning,  adaptability, and decision-making \cite{liu2017intelligence}. Yet, robot intelligence goes beyond cognitive abilities; it demands a deep understanding of commonsense and the ability to interact with users, translating human natural language into a logical sequence of robotic actions that are both physically feasible and coherent in the real world.

\begin{figure}[htp]
\centering
\includegraphics[width=0.475\textwidth]{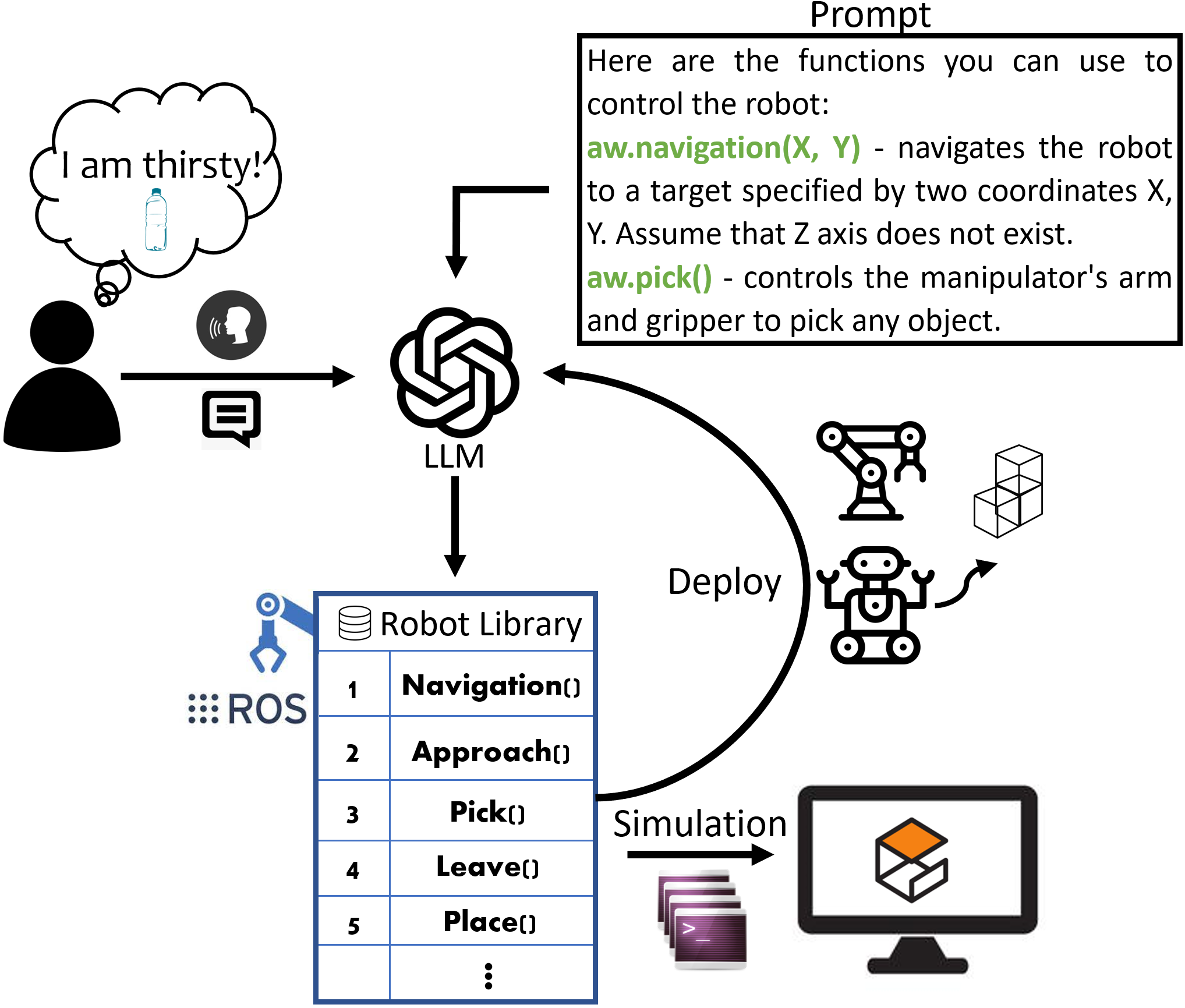}
\caption{RobotIQ: The overall framework of our approach.}
\label{fig:arch}
\end{figure}

Recent releases in high-performance Large Language Models (LLMs) have significantly impacted Natural Language Processing (NLP), empowering Human-Robot Interaction (HRI) and communication \cite{vaswani2017attention, ChatGPT}. Leveraging large training models and incorporating them into robotics, LLMs possess a close-to-human capability to understand context, reason through information, and generate coherent responses. However, while LLMs are designed to understand, generate, and process human language, they often lack true comprehension of commonsense or real-world knowledge, leading to potentially illogical or even biased outputs based on their training data. Prior to any action, the fundamental aspect of any robotics system is its ability to perceive and sense the world. Thus, naturally, the question that arises is ``How can we convert natural language into commands that can be understood and executed by robots?"

Robotic cognitive capabilities often require diverse learning approaches to optimize plan generation. Supervised learning methods employ explicit input-output pairs for specific tasks, ensuring precise alignment with predefined objectives. On the other hand, unsupervised learning facilitates a more generalized understanding of language and task structures, enabling adaptability to a broader range of scenarios.

Concerning robot intelligence, both supervised and unsupervised learning can be fine-tuned by prompting LLMs via APIs to serve as a bridge between the natural language comprehension of these models and the nature of symbolic AI. 
Creating a prompt for robotic task plans involves two components: the prompting functions and a strategy for searching answers \cite{liu2023pre}. Prompting functions outline the spectrum of actions a robot can perform, while the answer search process converts user input states into textual prompts, generating outputs—be it text-based or code-based—to facilitate the prediction of task plans. Towards this direction, APIs can facilitate the integration of symbolic reasoning and commonsense knowledge into LLM operations, allowing users to input instructions in an AI-friendly format for precise and controlled robot actions. In the context of task planning, this symbiotic relationship enables LLMs to utilize both their language comprehension abilities and the structured, rule-based reasoning characteristic of symbolic AI to generate comprehensive plans that align with specific objectives while also being adaptable to diverse scenarios \cite{huang2023voxposer, liang2023code, huang2023grounded, huang2021generalization, zhou2023isr}. In this study, structured symbolic AI is defined as the organization of knowledge through well-defined rules and representations, which enables systematic reasoning and logical inference.

Towards this direction, this paper aims to explore robot task planning with LLMs incorporating diverse learning approaches to both optimize plan generation and introduce a form of IQ into robotics, referred to as RobotIQ. Our goal is to sufficiently instruct LLMs to effectively utilize advanced targeted robotic cognition capabilities through API functions to facilitate robotic decision-making and task execution. The RobotIQ design
has been extensively tested in both simulated and real-world experiments, providing human operators with an AI-assistive robotic tool with navigation, object localization, fetch, and place capabilities based on human's spoken or written language. 

\subsection{Paper Outline}
The rest of the work is organized as follows: Section \ref{sec:RelWork} presents a thorough literature overview of the relatively recent LLM works and their integration into robotics, Section \ref{sec:robot_library} introduces the proposed robot library suite, 
elaborating on the methodologies and design of each module, while performing necessary evaluations, as needed, Section \ref{sec:Natural_language} describes the process of transforming natural language text or voice commands into executable code following prompt engineering principles, Section \ref{sec:EvalResults} presents the results of the simulated evaluation of the overall framework, Section \ref{sec:RealResults} presents the integration of RobotIQ into a real robot platform and the results of the real-world experiments, and Section \ref{sec:conclusions} provides an overview of the research and delves into its results and consequences.



\color{black}

\section{Related Work}
\label{sec:RelWork}
On one hand, robotic agents are autonomous entities capable of planning, decision-making, and performing actions to achieve complex tasks. On the other hand, due to its AI nature, the abilities of large language models (LLMs) to respond to dynamic scenarios have enabled their incorporation into various robotic applications, resulting in LLMs-powered agents, where LLMs function as the brains of these agents. 

To effectively serve as the robot's brain, LLMs must have the ability to understand, generate, and interact with language that is grounded in a specific physical, social, or perceptual context of the environment in which they are operating. For example, when a person instructs a
robot to pick up a bottle, the LLM-powered agent must map the word ``bottle” to a particular set of percepts in its operational environment, and then produce a plan or policy that causes its end effector to create a stable grasp of the bottle and lift it. This process of learning the connections between percepts and actions is called \textit{grounded language} \cite{tellex2020robots}. This means the language is connected to real-world situations, environments, and tasks, enabling the LLM to interpret and respond to instructions, questions, or interactions in a way that is relevant to the physical world.




From the LLM's perspective, modules and strategies employed in autonomous agents for founded language understanding, require planning, reasoning, feedback, and memory integration alongside human-like logic, step-by-step planning, and future-oriented reasoning. When tasked with a specific description, LLMs can understand and generate precise plans without additional training \cite{huang2022language}.
On the other hand, reasoning involves using an adapted LLM as an AI model to predict future outcomes and explore different strategies for task completion \cite{hao2023reasoning}.  In open-loop systems, LLMs generate plans based on closed-loop methods that provide feedback to LLMs, enabling re-evaluation and updates to ensure accurate task completion \cite{jin2023alphablock}. Several approaches \cite{singh2023progprompt, huang2022inner, jin2023alphablock} integrate LLMs into closed-loop systems, where feedback from actions is used to reassess and modify plans as necessary. Depending on the LLMs' capabilities, various methods either fine-tune the models, propose improved prompting strategies, or utilize different modules to boost agent's performance. More specifically, Song et al. \cite{song2023llm} propose an LLM-Planner for a few-shot planning for embodied agents capable of dynamically re-planing based on environmental perception to produce more grounded plans. Singh et al. \cite{singh2023progprompt} introduced ProgPrompt, a LLM prompting scheme for robot task planning incorporating commonsense reasoning and code understanding via situated state feedback from the environment. Ding et al. \cite{dingtask}, introduces LLM-GROP, a generalized prompting approach for object rearrangement tasks in varying scene geometry while Liu et al. \cite{liu2024delta} proposes an LLM-driven task planning approach called DELTA which translates environmental topology into actionable grounding knowledge.

In real-world scenarios, while LLMs excel at following instructions, their use in physically grounded tasks necessitates adjustments due to their limited real-world knowledge. 
Recent studies have explored the ability of LLMs for few-shot high-level robotics task planning \cite{liang2023code, huang2022language, xu2018neural, srinivas2018universal, akakziagrounding, song2023llm} for HRI, focusing on manipulation and navigation. In terms of manipulation, LLMs enhance a robot's ability and adaptability, leading to more effective performance in tasks such as object recognition, grasping, picking, and placing. They process visual and spatial data to identify the best methods for interacting with various objects \cite{jin2023alphablock, ha2023scaling}. Regarding navigation, LLMs boost a robot’s ability to navigate through complex environments with precision and accuracy. They generate viable paths and trajectories while considering various environmental factors, which is particularly valuable in settings that require precise and adaptable navigation in dynamic environments \cite{rajvanshi2024saynav, dorbala2023can, huang2023visual}. These studies utilize structured prompts containing predefined functions and examples to direct the model's generated responses. However, crafting prompts may present challenges due to the absence of matched natural language instructions linked to specific executable plans or sequences of robot actions, especially in real-world conditions. Having this in mind, our approach bypasses traditional search methods, directly producing a plan using custom-developed APIs that clearly describe the intended function behavior based on the robot's configurations that incorporate conditional reasoning and error correction.


\color{black}
Similar to our approach, \cite{huang2022language}, employ LLMs to generate open-domain plans in symbolic AI. Their methodology involves several steps: firstly, choosing a comparable task from the prompt example; secondly, generating task plans in an open-ended manner (answer search); and finally, aligning 1:1 predictions directly with robot actions. Additionally, \cite{vemprala2023chatgpt}, generates open-domain plans for robots using GPT-4. In this work, planning proceeds based on a high-level function library that allows ChatGPT to adapt to different robotics tasks. However, action-matching relies on predefined API functions derived from a simulator or a framework, thereby limiting the capabilities of a robot based on a generated text that ensures that the action is admissible in specific scenarios and not applicable in the real world.

The Transformer structure, introduced by \cite{vaswani2017attention}, has revolutionized NLP and shown great potential in the field of robotics. It has been applied in various reinforcement learning-driven tasks \cite{chen2021decision, janner2021offline} and pattern recognition \cite{he2022masked}. Additionally, model approaches such as SayCan \cite{brohan2023can} focus on connecting language models to interpret natural language commands and compute a value function that ranks the output actions available within a robot-specific library. On the other hand, RT-1 \cite{brohan2022rt} follows an end-to-end strategy, learning how language commands directly map to low-level robotic actions without relying on intermediate high-level functions.

Ye et al. \cite{ye2023improved} proposed a robot control system called RoboGPT utilizing ChatGPT \cite{ChatGPT} to control a robotic arm manipulator. The distinguishing characteristic of their methodology lies in the incorporation of off-the-shelf functions for robot control adopting a user-on-the-loop architecture. They introduced an AI assistant that generates prompts, presents the prompts to human operators, and waits for further instructions to proceed to the executable plan, emphasizing on the trust level between humans and robots. Although their approach showcased an improved trust level between humans and robots, RoboGPT is based only on generic robotic functions, prioritizing the bidirectional communication between humans and robots.

Jin et al. \cite{jin2024robotgpt} introduced a learning framework for robotic manipulation utilizing ChatGPT. The paper proposes leveraging a simulation environment and a natural language-based robotic API to enhance ChatGPT's problem-solving capabilities. The objective is to enable ChatGPT, through a trained agent called RobotGPT, to absorb knowledge at the task-planning level and improve task success rates by keeping ChatGPT in the loop acting as a decision, evaluation, and corrector until their code runs successfully. Their strategy, however, explores ChatGPT-generated code as a task planner which sometimes may produce minor bugs or syntax errors that cannot be directly applicable to real robots, and even if it could, it cannot guarantee the stability and safety of the system.

In contrast to the existing methodologies, this paper aims to enhance the capabilities of mobile robots by equipping them with planning abilities akin to human-level thinking, without imposing any limitations. This endeavor seeks to bridge the gap between humans and robots by leveraging secure and precise instructions from any LLM guided by natural language. More specifically, our proposed framework presents a holistic approach to open-domain planning for mobile robots. It employs a multi-phase strategy that combines the standardization and adaptability of high-level functions with the precision of direct action mapping to low-level robotic actions. These robotic functions are enclosed in a library suite that serves as a centralized repository, ensuring code reliability and unlocking the challenges of AI-generated code drift while employing LLMs for generating robotic-related code that may render the robot in invalid states that could compromise the robot's structural integrity or the failure of the assigned task. With modularity, version control, and comprehensive documentation, RobotIQ ensures consistency, facilitates updates, and enables easy management of any robotic platform, thereby allowing LLMs to translate user objectives expressed in natural language format into a logical sequence of high-level function calls, as depicted in Fig. \ref{fig:arch}.

\subsection{Contributions}
In a nutshell, the novelty of this paper lies in a system-focused contribution, namely RobotIQ, a framework designed for empowering mobile robots with human-level planning while integrating the following characteristics and software solutions:



\begin{itemize}
    \item  A ROS-based API pipeline for robotic actions and AI integration with flexible LLM application,

    \item Development of well-defined API-wise control functions as a robotic library suite supporting modular control,
    
   
    \item Enhance human-robot interactions where users can communicate with any robot in a natural language format by \textit{text} or \textit{voice} commands, 

    \item An open-source and easy-to-use robotic library suite for easy integration and adaptation across a wide range of robotic platforms and applications,
    
    \item A Sim-to-Real transfer learning-based policy for a robot navigation case.

    \item Evaluation of the proposed system in simulated and real-world experiments.

\end{itemize}


The theoretical contribution of this paper lies in simplifying the integration of intelligence (IQ) into robotic platforms. RobotIQ, offers flexibility by allowing a single framework to be used across a diverse array of robotic platforms, including ground, aerial, and underwater robots, as well as robotic arms while focusing on navigation, manipulation, perception, localization, and human-robot interaction presenting the most important capabilities in the field of robotics. 

A standout feature of our research is the development of a Sim-to-Real transfer learning-based policy for robot navigation, which represents a significant advancement in bridging the gap between simulation and real-world applications. This novel approach allows our system to effectively transfer learned behaviors from simulated environments to actual robotic operations, ensuring that navigation policies are both accurate and adaptable to real-world conditions. By leveraging this policy, our framework enhances the efficiency and reliability of robot navigation, setting a new benchmark in the field. This contribution not only addresses common challenges in transferring simulated knowledge but also demonstrates a substantial leap forward in improving real-world robotic performance.

Furthermore, built upon ROS, RobotIQ benefits from the robust tools and resources that ROS offers, ensuring compatibility with different hardware and software components and making it easier to deploy and extend the system across multiple robotic applications, leveraging the inherent advantages of ROS, such as its modularity, interoperability, and extensive community support. 

Our framework is designed to be compatible with any LLM, allowing it to integrate with a wide range of models and stay current with the latest AI advancements. This versatility enables users to customize the system based on specific needs, leveraging the unique strengths of different LLMs across different projects and industries, optimizing performance and resource usage by selecting the most appropriate model for each scenario. The ability to switch between LLMs also enhances the system's robustness, reducing the risk of underperformance in certain scenarios, and avoiding vendor lock-in ensures long-term sustainability and freedom in choosing technological solutions.

This universality is essential in an era where robotics applications are used in multiple industries, from healthcare and agriculture to manufacturing and exploration. By integrating advanced intelligence and NLP, robots can interact more effectively with humans, understand complex instructions, and adapt their behavior to dynamic environments, where they can interpret and act upon spoken commands in chaotic settings. The novelty of such a system lies in its holistic nature and in its departure from traditional approaches that rely on pre-defined API functions and are often limited to specific simulators or robots. Instead, an adaptable system like this provides a modular, customizable solution that extends beyond these constraints, offering a versatile tool for researchers and practitioners. This innovation not only enhances the efficiency and effectiveness of robotic systems but also accelerates the development of tailored solutions for various applications, leading to broader societal impacts. By addressing the need for a unified yet adaptable framework, this research supports the advancement of intelligent machines capable of making significant contributions to multiple industries, ultimately improving quality of life and driving progress on a global scale. To the best of our knowledge, RobotIQ is the first framework to (i) offer adaptability across various robotic platforms, (ii) support any machine learning, AI-based, and traditional approaches and translate them into API functions, and (iii) enable any LLM-based translation of natural language into programming code, all within a modular, public available and open-source framework\footnote{\url{https://github.com/emmarapt/RobotIQ.git}}.



\section{Robot Library Suite}
\label{sec:robot_library}
\subsection{Assumptions}
In the development of this research, several foundational assumptions were made to guide the direction and scope of the study. These assumptions include:
\begin{itemize}
    \item \textit{Structured Operating Environment}: The robot operates in a well-structured environment with minimal dynamic, unpredictable obstacles, ensuring effective performance of the reinforcement learning-based navigation.
    
    \item \textit{Sensor Accuracy and Reliability}: The robot's sensors, including LiDAR, cameras, and inertial measurement units, provide accurate and reliable data for localization, mapping, and object detection tasks.

    \item \textit{Clear User Commands}: User commands, delivered through text or voice, are assumed to be clear and unambiguous, enabling accurate translation into executable actions by the language model.

    \item \textit{Computational Capacity}: The robot's hardware possesses sufficient computational power to process data and execute algorithms in real-time.

    \item \textit{ROS Integration and Hardware Compatibility}: The availability of a ROS-compatible robotic hardware platform is assumed, which is essential for deploying the RobotIQ system effectively.
\end{itemize}

\subsection{Approach \& Rationale}
Building upon these foundations, the first step is to create a library of functions, which will subsequently used from the LLMs as stepping stones in drafting precise robotic plans based on the user's objective, as shown in Fig. \ref{fig:arch}. The motivation for selecting the following approaches for each module lies in our intention to demonstrate that the custom and modular ROS library of RobotIQ can support various machine learning and AI-based methods in both simulated and real-world environments. Our objective was to develop a system that is practical for real-world applications and compatible with a wide range of robots and tasks, ranging from motion planning to object localization, and grasping. To this end, we proceed with the development of well-defined API-wise control functions, either by leveraging open-source resources or engineering custom solutions, aimed at optimizing automation and human-robot interaction. The following subsections present the developed functions within the robot library suite. Having in mind that each module within the library can function either independently in a decentralized manner or within a centralized system, we outline each function's design and conduct necessary evaluations, as needed.


\subsection{Reinforcement Learning-Driven Navigation}
\label{sec::navigation}
In this section, we present a Reinforcement Learning (RL) setup employed to orchestrate the navigational function of a mobile robot.

The rationale behind employing a reinforcement learning controller is to empower the robot with the ability to dynamically explore any unknown environment and adapt its navigation strategy in real-time. Given that RobotIQ can support  a range of machine learning and AI methods, RL serving as a baseline, allows agents to learn optimal behaviors through interactions with their environment, which is particularly beneficial for complex decision-making tasks, highlighting RobotIQ's versatility in integrating self-improving capabilities in its robotic library suite.

More specifically, in our case, the primary goal of this function is to enable the robot to reach a target in an unknown environment while avoiding any non-traversable objects. To achieve this objective, we utilized the OpenAI gym \cite{brockman2016openai} framework to provide a standardized framework for benchmarking and evaluating the robot's performance. Additionally, we leveraged the capabilities of ROS \cite{quigley2009ros} and the Gazebo 3D simulator \cite{koenig2004design} to model and simulate the robot’s behavior within a virtual environment.

\subsubsection{Problem formulation}
Consider a scenario where a mobile robot must navigate from an initial position $r_p^{0}$ to a goal position $g$ on the $\mathbb{R}^2$ plane, surrounded by obstacles. For the purpose of developing a motion planning function for mobile ground robots, in this paper, we consider a nonholonomic\footnote{Non-holonomic mobile robots restrict lateral sliding} mobile robot. The robot's configurations at time $t$ are defined as $\xi(t) = [x(t)^T\, \, \theta(t)]^T = [r_{p_x}(t)\, \, r_{p_y}(t)\, \, r_h(t)]^T \in \mathbb{R}^3$, where $r_{p_x}$ and $r_{p_y}$ denote the coordinates of the robot and $r_h$ the robot's heading angle, both with respect to the $x_0$-axis of the global frame. The unicycle kinematic model of the robot is given by:

\begin{equation}
\dot{x} = v \cos(\theta); \quad \dot{y} = v \sin(\theta); \quad \dot{\theta} = \omega;
\end{equation}

where $v$ is the linear velocity and $\omega$ is the angular velocity of the robot. Therefore, the vector of control input is defined as $ \mathbf{u} = [v\, \,\omega]^T \in \mathbb{R}^2$. Assuming that the robot knows its initial configuration $\xi(0)$, the quaternion representation of $q$ based on the robot's heading is defined as $q= [q_x\, \, q_y\, \, q_z\, \, q_w]^T = [0\, \, 0\, \, \sin{\frac{r_h}{2}}\, \, 1]$.

Since our robot is moving only in the $x - y$ plane, by converting the quaternion representation into Euler angles, the heading of the robot is determined as:

\begin{equation}
    r_h = \tan^{-1}\left(2(q_w \times q_z + q_x \times q_y), 1 - 2(q_y^2 + q_z^2)\right)
\end{equation}

Knowing in advance the goal position ($x_g, y_g$), the angle between the robot's current position and the goal position in the x-y plane is defined as:

\begin{equation}
\theta_{goal} = \tan^{-1}\left(\frac{y_g - r_{p_y}}{x_g - r_{p_x}}\right)
\end{equation}

Finally, the equation for calculating the angular difference (also known as the angular error or heading error) between the desired orientation $\theta_g$ and the current orientation $r_h$ of the robot is given by:

\begin{equation}
    \phi = \theta_{g} - r_h
\end{equation}

The value of $\phi$ indicates how much the robot needs to rotate in order to align itself with the desired goal angle. Positive values indicate a counterclockwise rotation and negative values indicate a clockwise rotation. 

The robot should adjust its orientation by applying control actions to minimize this heading error and reach the desired orientation. Conceptually, framing the problem at the task-level and formalizing it as a Markov Decision Process (MDP), the goal of this function is to learn a policy $\pi(s)$ for the robot that maximizes heading to the goal ($\pi(s) = \omega$) while ensuring collision-free motions, defined as:

\begin{equation}
\begin{aligned}
    &  \pi^* = \arg\max_\omega \mathbb{E}\left[\sum_{t=0}^T \gamma^t r(s_t, a_t)\right] 
\end{aligned}
\label{optimization_eq}
\end{equation}

where $\gamma$ is the discount factor satisfying $0 \leq \gamma \leq 1$, $s_t, a_t$ are the state and action space at each time step, respectively, and $r(s_t, a_t)$ denotes the reward (Eq. \ref{r(s_t,a_t)}) indicating the outcome of a specific state-action pair. Having a constant linear velocity $v$, the objective is to compile all this information and determine an angular velocity $\omega_{t} \in \mathbf{u}$ that provides a complete alignment of the robot towards the desired 2-dimensional target position $x_{\text{target}}, y_{\text{target}}$.

\begin{figure*}[htp]
\centering
   \subfloat[]{\label{fig::rviz}
      \includegraphics[width=.30\textwidth]{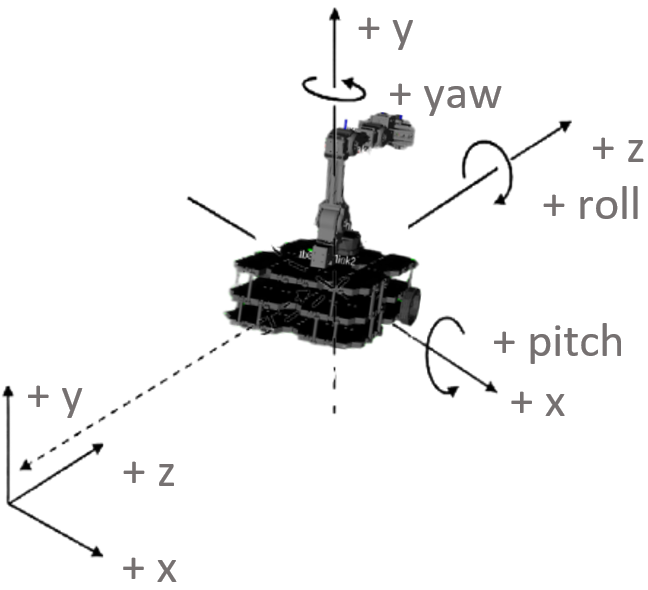}}
   \subfloat[]{\label{fig::rviz}
      \includegraphics[width=.70\textwidth]{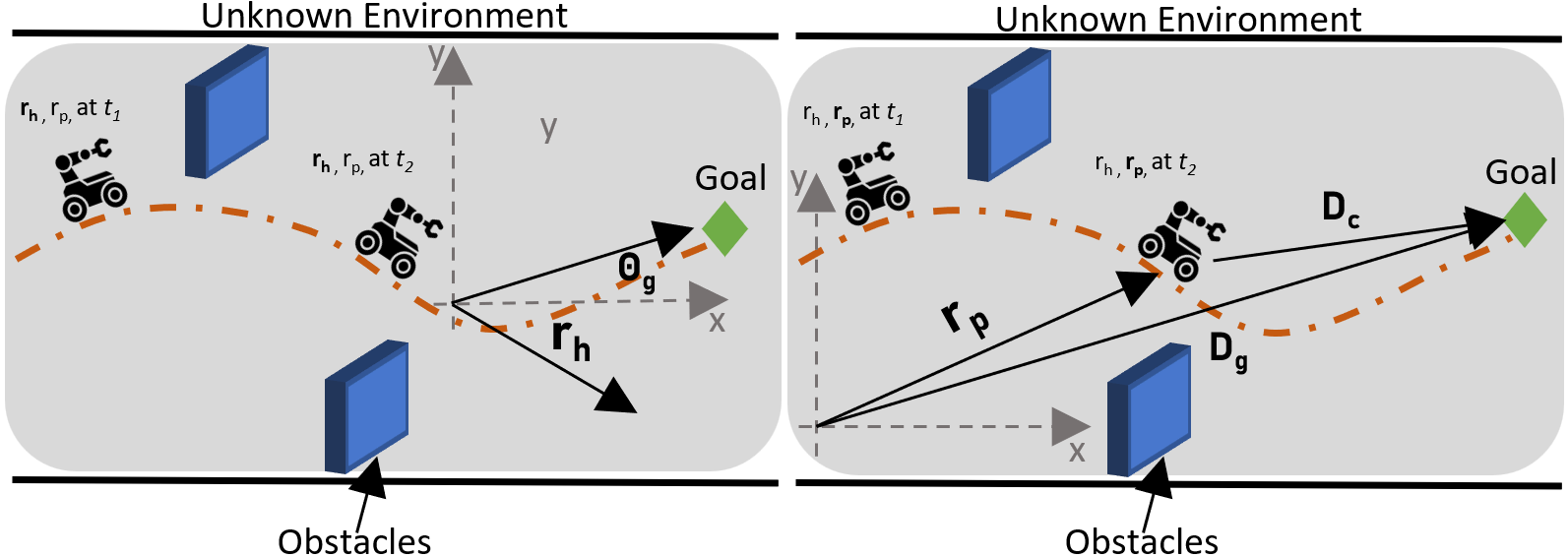}}
   \caption{Definitions of the state variables in an overhead view of the robot.}\label{fig:state_variables}
\end{figure*}

\subsubsection{State Space}
The robot is capable of acquiring extensive spatial information about its surroundings, constituting the state space. This information is obtained through a lidar-based sensing model that allows the robot to collect sampled data points in a fixed angle distribution, ranging from $-90$\textdegree to $90$\textdegree degrees while avoiding any possible collisions with the robotic arm, as depicted in the blue-shaded region of Fig. \ref{fig:rl_setup}.
The angular positions are discretized into a set of $n$ samples, each separated by a resolution of $\delta$, resulting in a uniform distribution falling within a scanning range $R_{min}$ to $R_{max}$. As a result, the state space can be represented as a multidimensional state space where the boundaries for each dimension of the observation space are defined as lower bound ($l_j$) and upper bound ($h_j$), as follows:

\begin{equation}
\begin{aligned}
l_j = \begin{cases}
    R_{min}, & \text{for } 1 \leq j \leq n \\
    -\pi, & \text{for } j = n + 1 \\
    0, & \text{for } j = n + 2,
\end{cases} \quad \forall j \in [1, n]
\end{aligned} 
\end{equation}

\begin{equation}
\begin{aligned}
h_j = \begin{cases}
    R_{max}, & \text{for } 1 \leq j \leq n \\
    \pi, & \text{for } j = n + 1 \\
    d_{max}, & \text{for } j = n + 2,
\end{cases} \quad \forall j \in [1, n]
\end{aligned}
\end{equation}

For the first $n$ dimensions, the lower bound is set as $R_{min}$, while the upper bound is set as $R_{max}$ ensuring that the scanning range of the sampling points falls within the defined range. The $(n + 1)_{th}$ dimension is assigned a lower bound of $-\pi$ and an upper bound of $\pi$, whereas the  $(n + 2)_{th}$ dimension has a lower bound of 0 and an upper bound of $d_{max}$, representing the minimum and the maximum distance from the goal, respectively. Consequently, at each time step, the mathematical representation of the state space $s_t$ can be defined as follows:

\begin{equation}
s_t = \{x \in \mathbb{R}^n \,|\, l \leq x \leq h \} 
\end{equation}

\subsubsection{Action Space}
Considering that the movement capabilities of the robot can be either discrete or continuous, the action space is defined accordingly. The robot's position in both cases is represented by the corresponding $x$, $y$ coordinates of the grid, i.e., $r_p(t)$ $=$ $[x(t), y(t)]$.

Let the action space be denoted by $A$. In case of discrete moving capabilities, the action space is defined as follows:

\begin{equation}
A = \left[ \left( \frac{N-1}{2} - a \right) \cdot \Delta \theta \right]_{a=0}^{N-1}
\end{equation}

where $N$ is the size of the possible actions, $a$ is the index of the action, and $\Delta \theta$ is the angular step. The angular step can be calculated as:

\begin{equation}
\Delta \theta = \frac{\omega_{max}}{(N-1)/2}
\end{equation}

where $\omega_{max}$ is the maximum angular velocity of the robot.  This formulation facilitates the calculation of the sequence of actions based on the action space's size and the robot's maximum angular velocity.

In case of continuous actions, the action space is defined as follows:

\begin{equation}
A = a \in \mathbb{R} \mid \omega_{min} \leq a \leq \omega_{max}
\end{equation}

\begin{figure}[htp]
\centering
\includegraphics[width=\columnwidth]{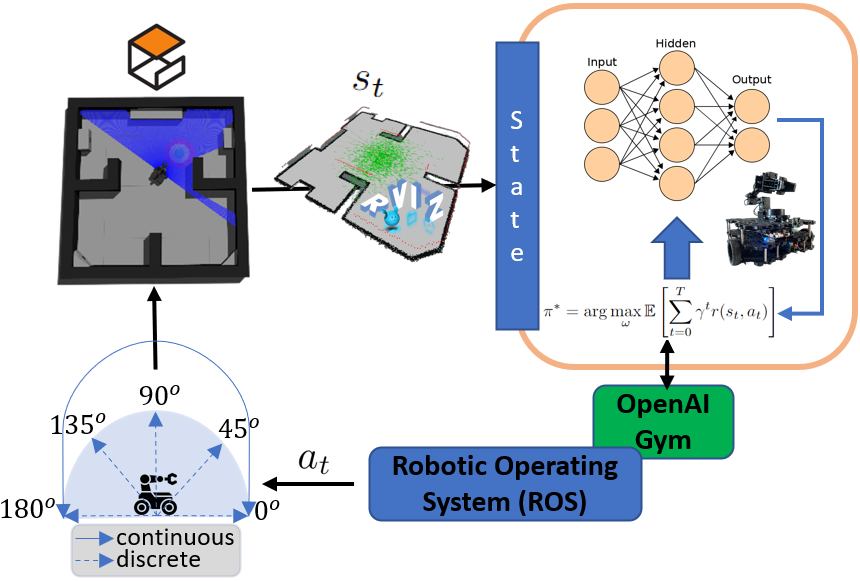}
\caption{Overview of the experimental architecture for reinforcement learning, depicting the alignment between the environment and its associated state-action pairs.}
\label{fig:rl_setup}
\end{figure}

In both cases, the 1-dimensional action of every time step includes the angular velocity of the mobile robot. Backward movements are not considered as the laser findings do not provide coverage for the rear region of the mobile robot. Fig. \ref{fig:rl_setup} presents an illustrative example of a registration between the environment and the corresponding state-action pair representation.

\subsubsection{Reward Function} 
Rewards are numerical values that represent the desirability or quality of a particular state-action pair. Having in mind that the ultimate goal is for the robot to reach a target in an unknown environment, the reward for solving Eq. \ref{optimization_eq} at each timestep $t$, is defined by the angle $r_h$ and the distance $r_p$ of the robot towards the target as follows:

\begin{equation}
    r(s_t, a_t) = (5 \cdot  r_{\text{{yaw}}(t)}) \cdot 2^{\frac{D_g}{D_c}}
    \label{r(s_t,a_t)}
\end{equation}

where the $r_{\text{{yaw}}(t)} = 1 - \phi(t)$ is a reward value between 0 and 1, representing the agent's alignment with the desired heading while $2^{\frac{D_g}{D_c}}$ represents a value that quantifies the relationship between the current distance $D_c$ and the goal. It measures how far the robot is from its goal relative to the maximum distance and provides a ratio that can be used to assess the progress or proximity of the robot to its destination.

By increasing the yaw reward by a factor of 5, it undergoes a proportional amplification, extending its range from 0 to 5. This rescaling operation imparts greater potency to the yaw reward's impact on the ultimate reward outcome. Additionally, when the rounded and re-scaled yaw reward is multiplied by $2^{\frac{D_g}{D_c}}$, it intensifies the reward magnitude as the agent approaches the desired target. Note that this factor is adjustable, and exploring alternative values can help fine-tune the reward function for various scenarios or performance objectives.

The action space, either discrete or continuous, encompasses potential movements that may render the robot in invalid states, such as colliding with an obstacle that could compromise the robot's structural integrity. Therefore, the robot must be able to recognize and actively avoid these unfavorable movements. Towards this direction, an additional penalty denoted as $r_{collide}$ = $-q \in \mathbb{R}$ is introduced when the subsequent movement of the robot leads to an invalid state. Conversely, a reward of $r_{goal}$ = $+q$ is given to the robot upon successful attainment of its target. For both cases, the penalty or bonus reward marks the completion of the current episode, prompting the initiation of a new one.

Putting everything together, the reward in each time-step is defined as follows:

\begin{equation}
    r(t) = \begin{cases}
        r_{collision} , & \textit{collision} \\
         r_{goal}, & \textit{goal} \\
        (5 \cdot  r_{\text{{yaw}}(t)}) \cdot 2^{\frac{D_g}{D_c}} , & \textit{otherwise} \\
    \end{cases}
    \label{reward_eq}
\end{equation}

\subsubsection{Performance Evaluation}
Having defined the fundamental attributes that constitute a reinforcement learning-driven mobile robot navigation function, in this subsection, we proceed with an experimental evaluation of the target-reached environment utilizing the Gazebo framework. The study focused on several model-free Reinforcement Learning (RL) algorithms, namely PPO \cite{schulman2017proximal}, TRPO \cite{schulman2015trust}, VPG \cite{yuan2022general}, DDPG \cite{ddpg} and TD3 \cite{dankwa2019twin}. Each experiment consisted of the agent interacting with the environment for 250 epochs, with each epoch comprising 3000 interactions with the environment.

\begin{figure}[htp]
\centering
   \subfloat[Discrete action space.]{\label{fig::rl_perfomance_discrete}
      \includegraphics[width=0.90\columnwidth]{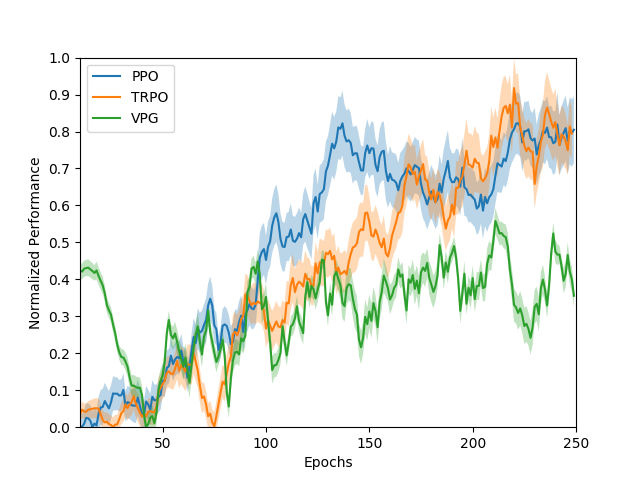}}
      
   \subfloat[Continuous action space.]{\label{fig::rl_perfomance_continuous}
      \includegraphics[width=0.90\columnwidth]{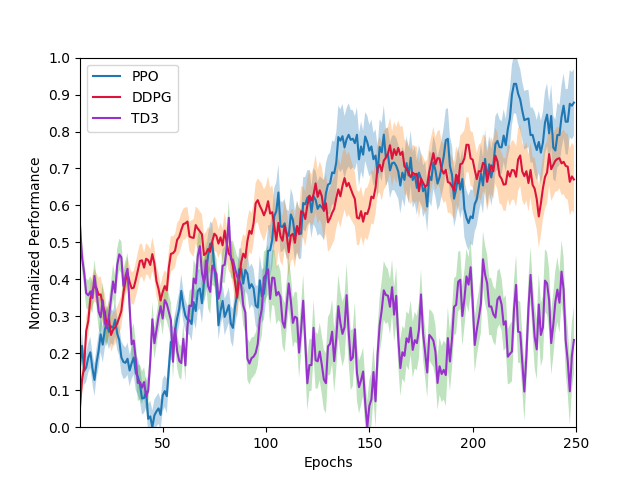}}

   \caption{Learning curves for the simulated environment with discrete \& continuous action space.}\label{fig::rl_perfomance}
\end{figure}

Fig. \ref{fig::rl_perfomance} illustrates a comprehensive comparative analysis of the aforementioned approaches. Each RL agent is represented by bold colored lines depicting the total rewards achieved in episodes, while the encompassing transparent surfaces depict the standard deviation. Additionally, the episode's rewards (scores) are normalized to a range between 0 and 1, where 0 denotes an initial invalid action taken by the robot ($r_{collision}$ in equation \ref{reward_eq}), and 1 corresponds to the theoretical maximum reward achievable ($r_{goal}$ in equation \ref{reward_eq}).



A clear-cut outcome is that the PPO algorithm achieves the highest average episodic reward in both discrete and continuous action spaces.

\subsection{Augmented Reality-Driven Object Localization}
In this section, we introduce an Augmented Reality (AR) pose estimation framework built upon ROS for detecting and tracking fiducial markers positioned beneath a location intended for object pickup or placement. Termed as ``Approach()", the primary objective of this function is to localize and reveal these locations and orchestrate precise navigational instructions to the robot, directing it toward them with utmost accuracy. This step is crucial to plan and execute autonomous manipulation tasks based on visual tracking, especially in scenarios where the camera and the robotic arm do not share the same reference frame. In robots equipped with cameras, this can be achieved by adopting a computer vision algorithm to detect visual markers located in the desired locations. To this end, for the AR pose estimation we utilized the $ar\_track\_alvar$ package standing as the prevalent choice providing comprehensive functionalities for marker detection, pose estimation, and tracking \cite{ar_track_alvar}.  As to the fiducial markers, ArUco markers were selected as they exhibit rapid recognition capabilities and a minimal level of ambiguity specifically in indoor environments \cite{garrido2014automatic}. Through the determination of the robot's camera position and orientation, we establish the desired pose of the locations for object pickup or placement, as shown in Fig. \ref{fig::rviz}.

Once the marker is detected, the $ar\_track\_alvar$ package continues to track the markers in subsequent frames. This tracking process involves predicting the marker's location by using information from previous frames to refine the pose estimation. By establishing a continuous estimation of the object's pose, comprising both its position and orientation, we subsequently disseminate highly accurate linear ($v$) and angular ($\omega$) velocities to the robot's designated topics, along with a user-defined parameter $x$ indicating the distance from the object. 
This enables the robot to swiftly approach, align, and orient itself toward the recognized marker, achieving a designated proximity of $x$ meters from the object. Additionally, the ``Leave()" function, characterized by the same definition, incorporates the introduction of a user-defined parameter denoted as $x$, signifying the distance in meters from the target. Following these steps, we establish a complete solution for AR marker detection and tracking.

\subsection{Pick \& Place}
The choice of a specific grasp pose may strongly depend on the position of the arm. To this end. by utilizing the advancements of the continuous pose estimation enabled by AR marker detection, we have deployed a method for manipulation tasks based on the accurate alignment of the robot in relation to the object. For controlling the arm trajectory planning, we have utilized the MoveIt package \cite{sucan2012open}, particularly relying on the Open Motion Planning Library (OMPL). MoveIt represents a state-of-the-art open-source robot programming framework, renowned for its multiple functionalities tailored to manipulation tasks, encompassing kinematics, trajectory planning, collision avoidance, and grasping modules. As an integral package of ROS, MoveIt seamlessly integrates with the ROS ecosystem via the move$\_$group node, allowing for effortless transmission of motion planning requests to the robot's manipulator. These requests typically include specifying a target pose or trajectory that the robot needs to achieve. Subsequently, the move$\_$group node plans a collision-free trajectory planning process to reach the desired target, transmitting the corresponding commands to the robot's motion controller.

\begin{figure}[htp]
\centering
\includegraphics[width=\columnwidth]{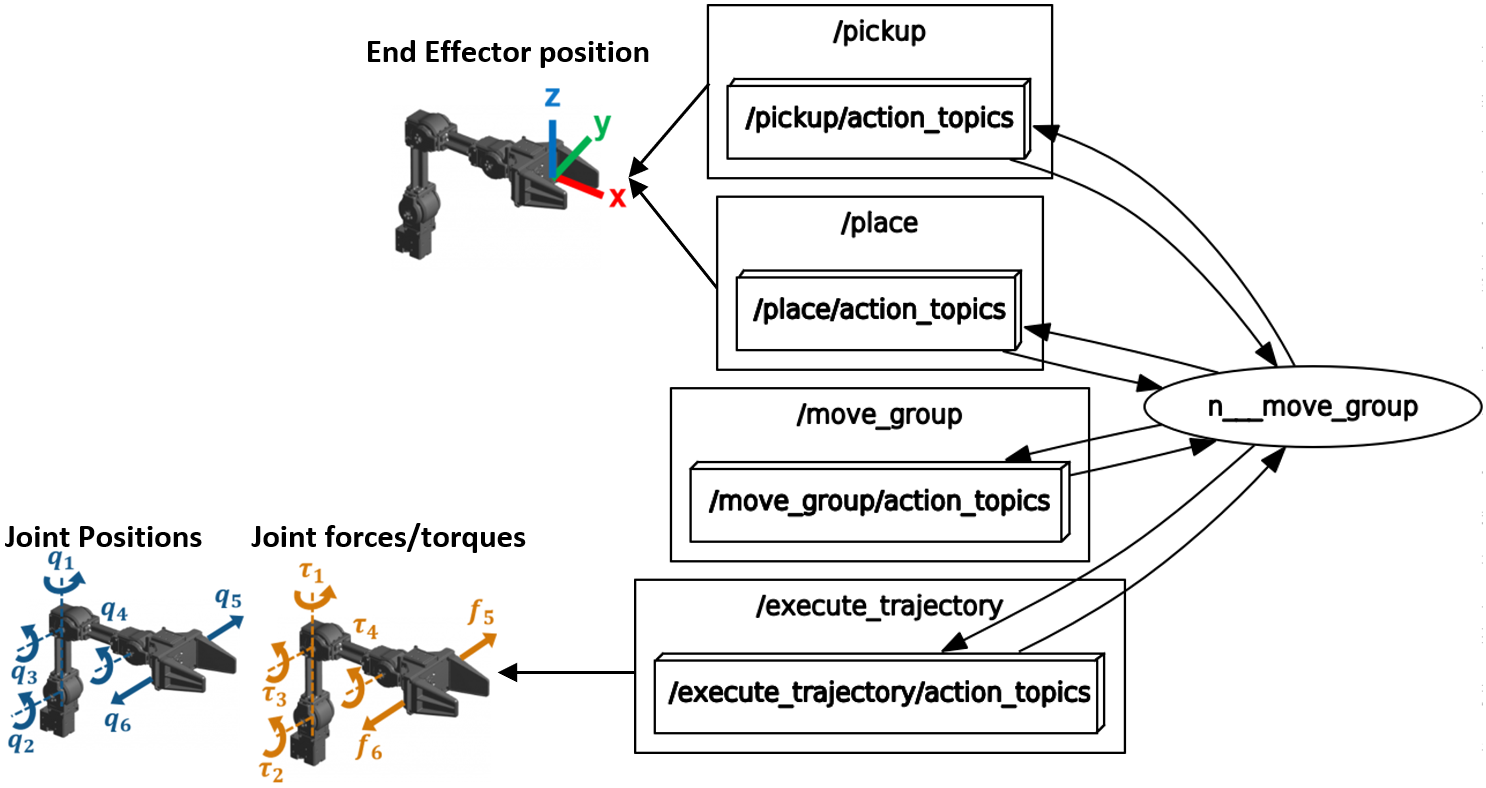}
\caption{
Illustration of move$\_$group node communicating with ROS topics, facilitating information exchange on joint forces, torques, and poses.}
\label{fig:moveit_contacts}
\end{figure}

\begin{figure*}[htp]
\centering
\includegraphics[width=\textwidth]{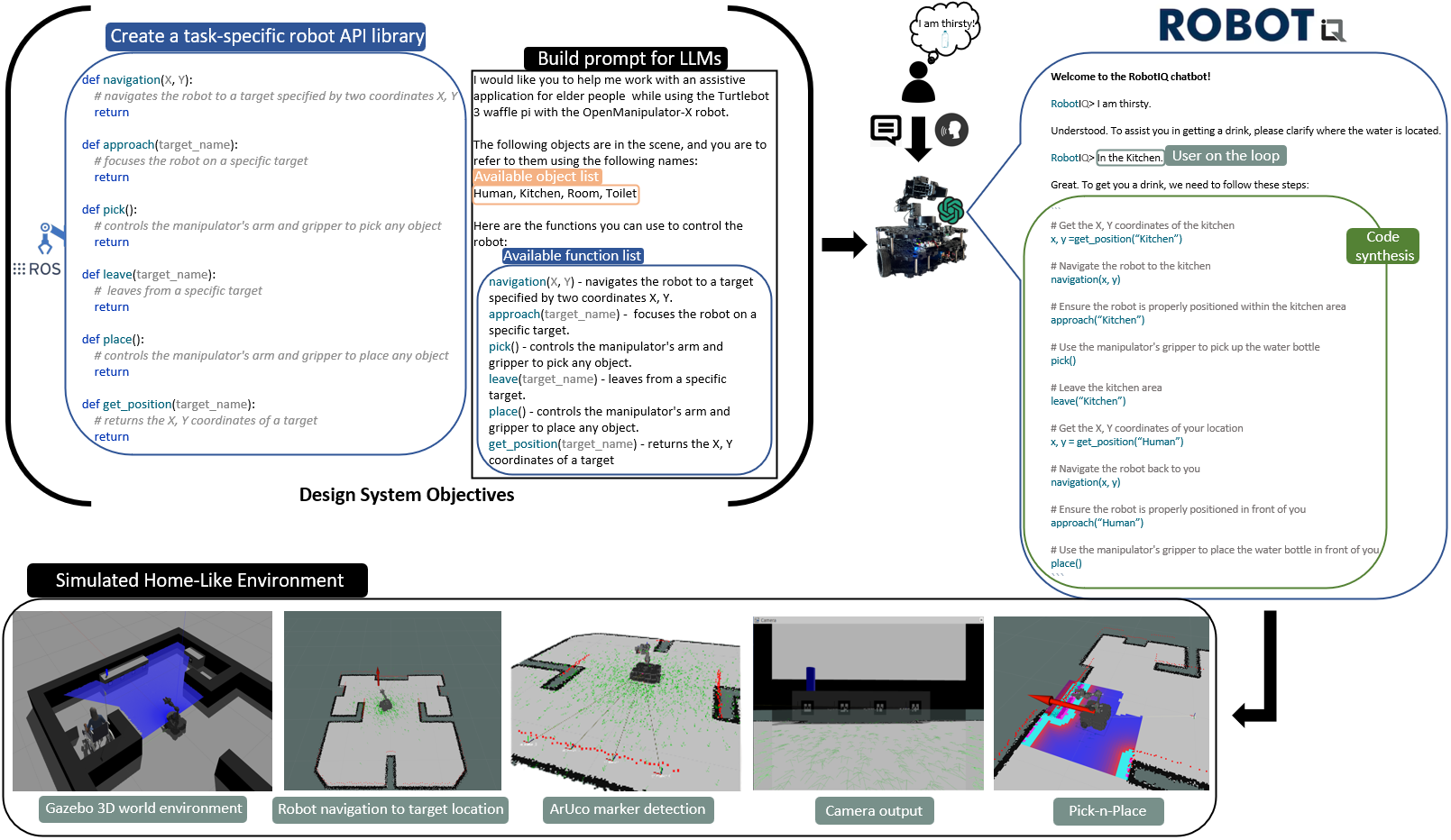}
\caption{
Figure 6 illustrates the process of transforming natural language commands into executable code using a suite of custom robotic library functions following prompt engineering principles. It shows how user inputs, given through free-form dialogue, are processed by the language model and converted into specific API calls. The main elements include modules for navigation, manipulation, localization, and human-robot interaction. The
navigation module handles movement commands, while the manipulation module deals with tasks like picking and placing objects. The localization module keeps track of the robot’s position, and the human-robot interaction module manages communication. This figure demonstrates the entire workflow from user command to robotic action, highlighting how each step seamlessly integrates to enable the robot to
interpret and execute tasks within the RobotIQ framework.}
\label{fig:prompt}
\end{figure*}

Fig. \ref{fig:moveit_contacts} illustrates a high-level system architecture diagram depicting the MoveIt move$\_$group node and its connectivity with the robot. In particular, the move$\_$group node establishes communication with the robot by publishing and subscribing to ROS topics, facilitating the exchange of information related to the equilibrium state of the arm's joint forces, torques, and poses to achieve its intended objectives. These topics serve as communication channels facilitating the exchange of information between different system components. Utilizing the robot's topics, the MoveIt move$\_$group node effectively communicates with the robot, enabling real-time feedback and status updates during motion planning and execution. As depicted in Fig. \ref{fig:moveit_contacts}, the intricate connections showcase the bidirectional data flow between the move$\_$group node and the robot's topics, showcasing the real-time exchange of joint state information, planning scene updates, and execution trajectories.

\section{Natural Language for Human-Robot Interaction}
\label{sec:Natural_language}
In this section, we establish a workflow that integrates LLMs with robotic control modules to build an intelligent AI robot control assistant akin to human-level planning in a \textit{zero-shot} fashion. As illustrated in Fig. \ref{fig:arch}, the RobotIQ process initially transforms text or voice human commands into written input for the AI assistant to process. In this study, the AI assistant's decision-making core employs on GPT-4 \cite{achiam2023gpt} through the OpenAI API to process and respond to the provided information However, it must be noted that RobotIQ is not restricted to a specific LLM but rather is versatile and can be configured to work with any LLM.

By designing a clear and concise description of the desired robotic task objectives and their context (i.e., constraints and requirements), humans enable LLMs to generate natural responses resulting in a bidirectional prompt-based dialogue system that seamlessly aligns with diverse real-world robotic tasks and scenarios.

Hence, we have encapsulated the previously proposed robotic modules within high-level functions, effectively serving as user-friendly wrappers over the actual implementations. Typically, this is followed by a section of explanatory text and a subsequent block of code. A demonstrative example showcasing interactive communication between the user and GPT-4 model is illustrated in Fig. \ref{fig:prompt}. 

\color{black}
\section{Simulated Evaluation}
\label{sec:EvalResults}
To assess the whole performance of the proposed robotic system, custom-designed prompt-engineering principles, and the code generated by the LLM, we first developed a simulated experimental setup. The ultimate goal of this simulated evaluation was to extensively examine and assess both the performance of the custom robotic libraries one-by-one in isolation and the overall RobotIQ system as a whole, as well as the ability of the LLM to utilize predefined libraries to generate complex and detailed robotic plans until achieving a robust and ready-to-deploy code for the physical robot.

For all the simulated experiments, a virtual Turtlebot3 Waffle Pi equipped with the OpenManipulator-X manipulator (4-DoF) was utilized. For the representation and visualization of the trajectories we used the RViz (Robot Visualization) program \cite{Rviz}, an auxiliary tool for visualizing robot and data including RGB camera and LiDAR sensor outputs. The robot recorded the laser range findings from an LDS-02 lidar sensor characterized by a field of view (FOV) of 180\textdegree and an angular resolution of 1\textdegree while the scanning range was ranged from 0.120 to 3.5 $m$. The simulation was hosted in a common laptop (Intel Core i7-8750H CPU) with Ubuntu 16.04 LTS and a combination of v7.7/Kinetic as a version of Gazebo and ROS package, respectively. 

\subsection{Experimental Scenario: Robot-as-a-Service}
\label{ref:home_service_scenario}
The experimental evaluation revolved around a home service scenario, incorporating an assistive
application designed for elderly people, where the goal is to move different objects from one place to another with given rules and return to the starting point, all through a home-like environment.



It is important to clarify that the home service scenario was used as an illustrative example to demonstrate the effectiveness of RobotIQ. While this scenario was used for the evaluation, it was not the only possible application. The scenario was designed by the authors to cover all the key domains—such as navigation, manipulation, perception, localization, and human-robot interaction—that the system aims to evaluate. This means the same functionality and results would have been achieved with any other scenario involving these domains, highlighting RobotIQ’s versatility across a wide range of use cases.

Thus, in our study, to define these rules, we employed custom-designed prompt-engineering principles, and GPT-4 was utilized to translate natural language instructions 
into robotic actions using our custom robot library, as illustrated in Fig. \ref{fig:prompt}. The robot's primary objective was to perform a diverse set of tasks mirroring human abilities serving as a unique Robot-as-a-Service (RaaS) model and ensure that the assistive application would be highly effective and reliable, prioritizing safety for both the robot's integrity and its users in real-world scenarios. In particular, the scenario that was considered was the following: \textit{An elderly human is on a wheelchair. He/She is thirsty and needs water to hydrate. The bottle of water is located in the kitchen. The human informs the system about his/her needs using either voice or text commands. Based on human's commands, the robot must then perform a set of tasks to deliver the bottle of water, i.e.: `Go to the kitchen', `Pick the bottle of water', `Leave the kitchen area', `Navigate towards the human' and finally `Place the bottle of water near the human'.}

\subsection{Performance Analysis}
To assess the performance of our proposed system, we conducted two evaluations for the chatbot and voicebot interfaces: the first focused on evaluating the standalone performance of each system, and the second involved a comparative analysis between the two. 

\begin{figure*}[htp]
\centering
   \subfloat[Chatbot simulation results.]{\label{fig:chatbot_perfomance}
      \includegraphics[width=0.9\textwidth]{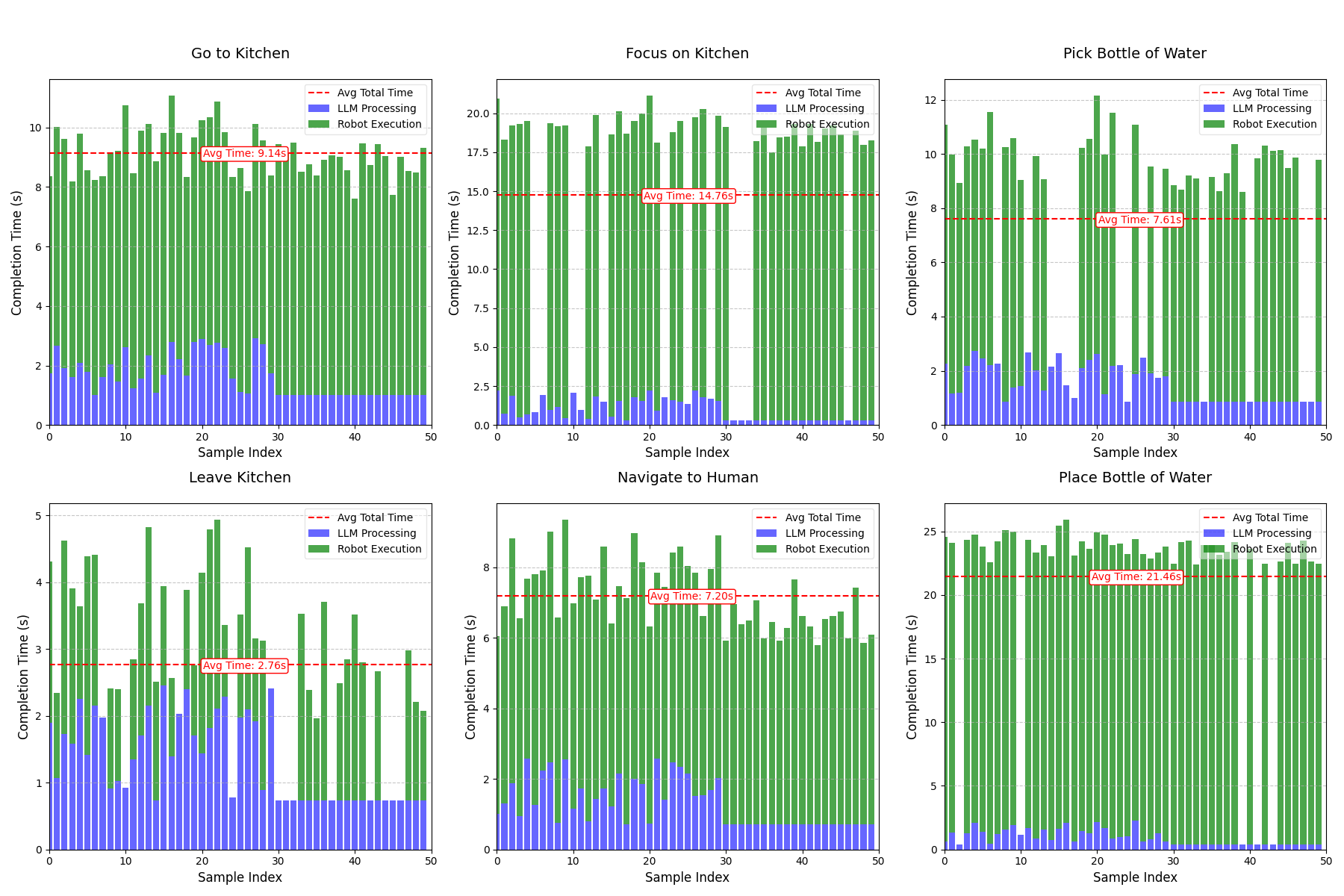}}

   \subfloat[Voicebot simulation results]{\label{fig:voicebot_perfomance}
      \includegraphics[width=0.9\textwidth]{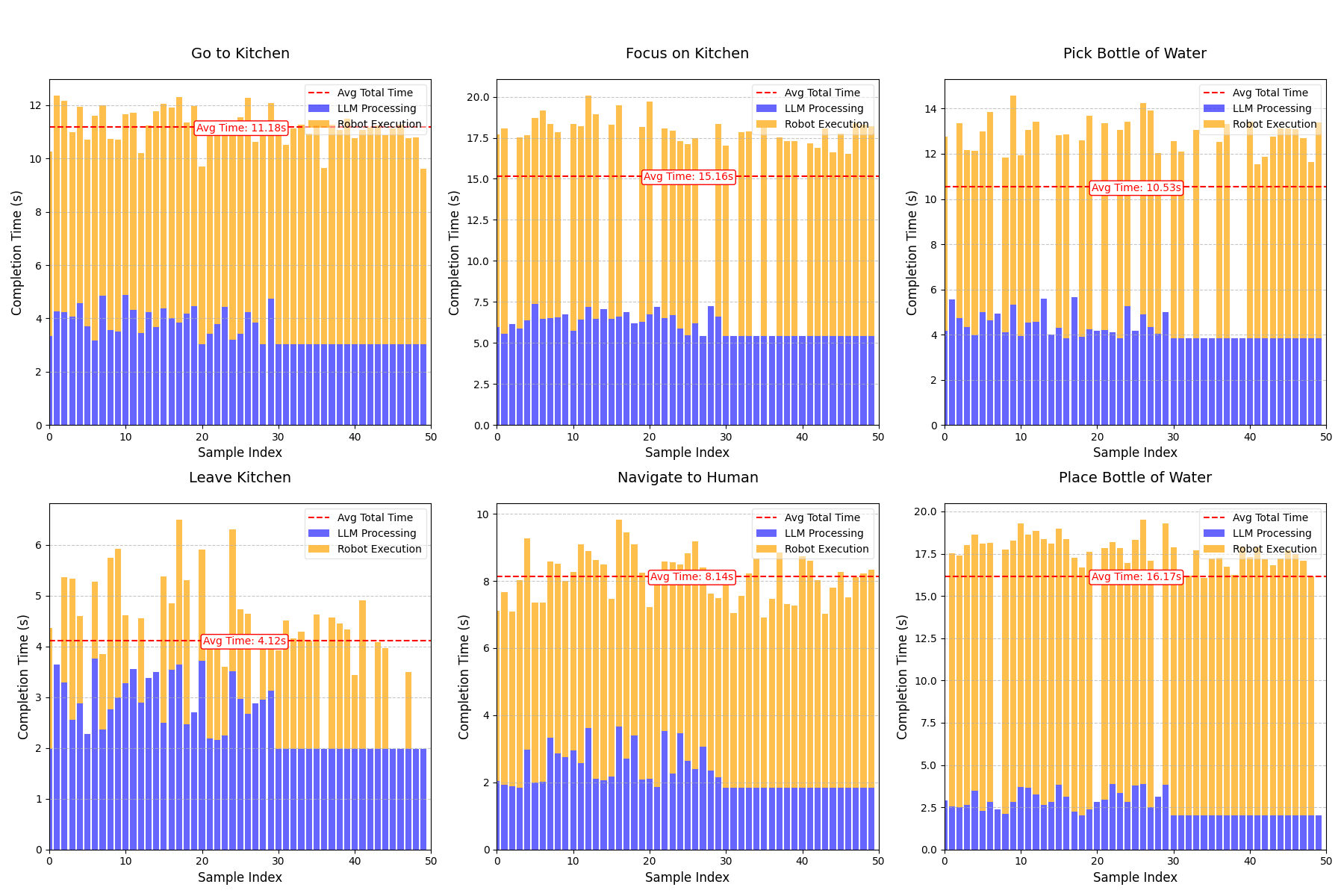}}

   \caption{Performance simulated evaluation between Chatbot and Voicebot. The plots show a distinct contribution from LLM processing (blue), robot execution (green), and average (red) times .}
   \label{fig:Chatbot_vs_Voicebot}
\end{figure*}

The first evaluation focuses on two critical time-consuming segments: \textit{LLM processing times} and \textit{robot execution times}, which together constitute the total task completion time ($T_{\text{Total}} = T_{\text{LLM}} + T_{\text{Robot}}$). This metric captures the overall efficiency of the system in executing natural language commands, allowing us to pinpoint delays introduced by either the decision-making process (LLM) or the physical execution of tasks by the robot. Additionally, we factor in the success rate of task execution, where failures or delays may occur due to incomplete or error task performance. This analysis offers insights into the efficiency and responsiveness of each interface in facilitating human-robot interaction. The results are visualized using bar plots to provide a comprehensive understanding of performance trends across the multiple tasks, as illustrated in Fig. \ref{fig:Chatbot_vs_Voicebot}.



      

\begin{figure*}[htp]
\centering
\includegraphics[width=0.9
\textwidth]{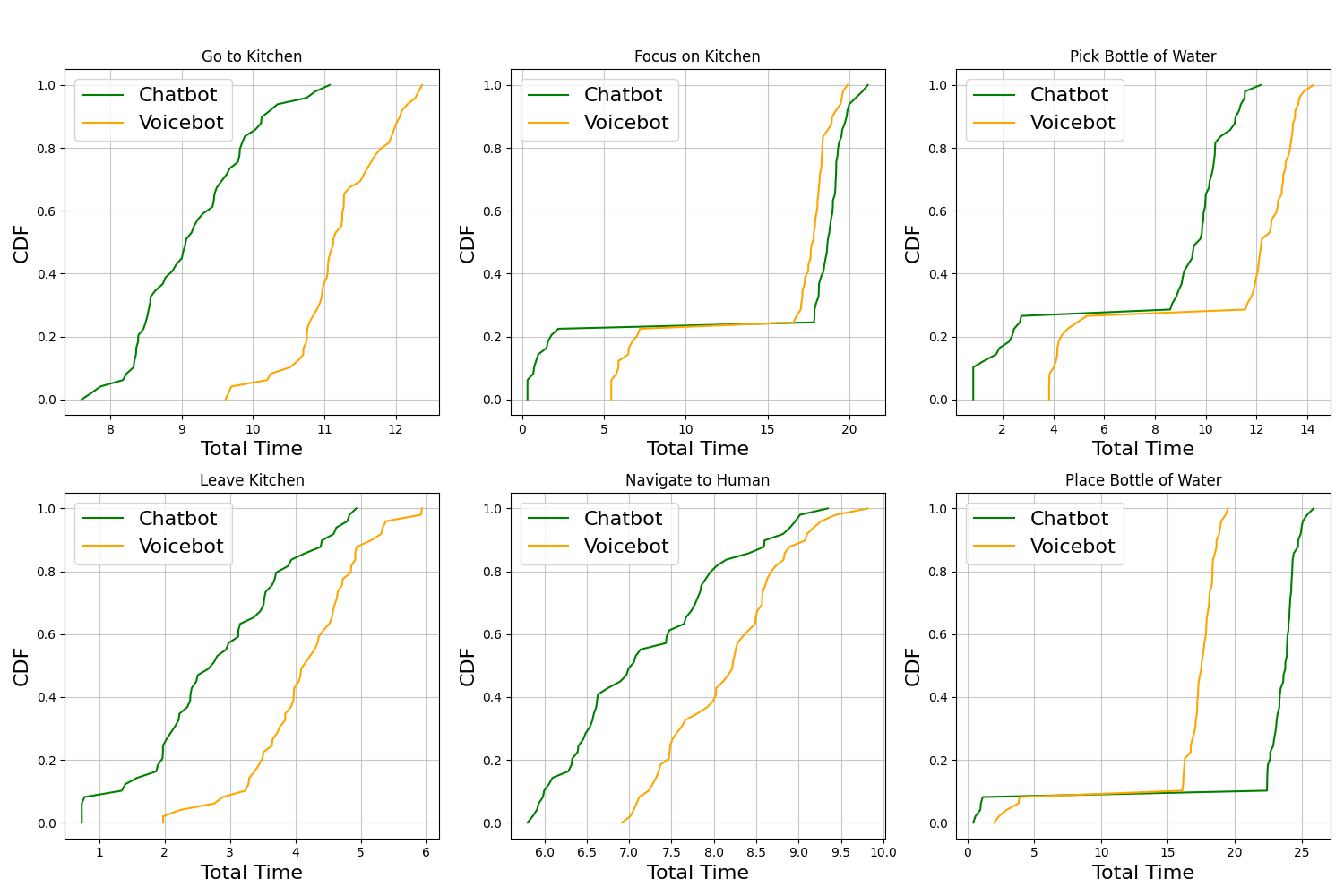}
\caption{Cumulative Distribution Function (CDF) between Chatbot and Voicebot. Each subplot corresponds to a specific task, with the chatbot results shown in green and the voicebot results in orange. For each system, the CDF curve provides a cumulative view of the proportion of tasks completed, 
allowing for a direct comparison of their performance profiles. Steeper curves reaching closer to 1 at lower times indicate better performance, as a larger fraction of tasks are completed more quickly.}
\label{fig:CDF}
\end{figure*}


To ensure robustness, we conducted 50 simulated experiments for each interface, evaluating both the prompt engineering principles applied to the LLM (hence the inclusion of LLM processing times in the Fig. \ref{fig:chatbot_perfomance} \& \ref{fig:voicebot_perfomance}) and the performance of developed functions. As to the prompts, the first 30 samples included designing, analyzing, and refining the inputs used to extract responses from the AI model. The best-performing prompt across the first 30 samples was then used consistently across both the chatbot and the voicebot, ensuring consistency in performance analysis under steady-state conditions. Notably, the prompt was identical for both interfaces, ensuring a fair comparison and consistent baseline for performance analysis. As to the developed functions, the tasks in which robot execution times exceeded a 30-second timeout were set to zero, simulating unsuccessful attempts.

Both the chatbot and voicebot systems demonstrate high reliability, with success rates exceeding 70\% across all tasks, as shown in Fig. \ref{fig:chatbot_perfomance} and Fig. \ref{fig:voicebot_perfomance}. Compared to the chatbot, the voicebot exhibits consistently higher LLM processing times due to the additional overhead of speech-to-text processing. For example, in tasks such as Focus on Kitchen and Pick Bottle of Water, the LLM processing times for the voicebot are approximately double those of the chatbot. This disparity highlights the computational burden of handling voice-based interactions. However, the robot execution times for the voicebot align closely with those of the chatbot, indicating that the physical execution of tasks is interface-agnostic. Despite the differences in interaction modalities, the average task completion times for both systems remain comparable, with the added overhead of the voicebot's LLM processing having minimal impact on overall performance, underscoring their robustness in interpreting and executing natural language commands.


For the second evaluation, to analyze the comparative performance of these systems, we use cumulative distribution function (CDF) plots \cite{pianosi2015simple}. Figure \ref{fig:CDF} provides a direct comparison of the total task completion times for the chatbot and voicebot systems.  By plotting the CDFs for the chatbot and voicebot systems, we aim to illustrate the distribution and variability of their execution times, thereby gaining a deeper understanding of their performance during the simulation experiments. This comparison highlights an important trade-off: while the voicebot offers a more natural and intuitive human-robot interaction, it incurs additional processing delays that may not be suitable for time-sensitive applications. On the other hand, the chatbot's text-based interaction provides faster responses, making it better suited for scenarios requiring fast task execution.
\color{black}


The overall outcome of the conducted simulated experiments for both chatbot and voicebot underscores the potential of the Robot-as-a-Service (RaaS) model, featuring the proposed system as a comprehensive and supportive home service solution for elderly people. By incorporating human feedback and domain-specific knowledge into the prompt-engineering process, the LLM-driven code developed for the robot exhibits an impressive level of accuracy and robustness, enabling seamless interaction with our custom suite robot library. By leveraging AR markers, precise object detection and pose estimation, autonomous navigation, and proficient object manipulation, whether it is picking or placing, RobotIQ showcased exemplary dexterity and adaptability over LLM's commands. 

A video demonstration showcasing the simulation experiments involving the aforementioned robotic services over the ChatBot and the VoiceBot are attached to the following links\footnote{\url{https://youtu.be/RrMleHW5YRw}}$^,$\footnote{\url{https://youtu.be/z3siFzY07w8}}.

\section{Real-world Experiments}
\label{sec:RealResults}
Having successfully demonstrated the proposed robotic system in the simulated experiments, we next proceed to conduct the same scenario in the real world. For the real-life experiments, comparable to the simulations, a Turtlebot3 Waffle Pi equipped with the OpenManipulator-X manipulator, an RGB camera, and a LiDAR sensor were utilized. In this case, the Turtlebot3 Waffle Pi was powered by a Raspberry Pi 4b model as its onboard computer, and the operating system chosen for this setup was Rasbian Buster. To facilitate seamless communication, the Turtlebot3 connected wirelessly to a remote PC powered by an Intel Core i7-8750H CPU while running Ubuntu 16.04 LTS and ROS kinetic. The communication between the Turtlebot3 and the remote PC was established using the ROS middleware, with the ROS\_MASTER\_URI set to the IP address of the remote PC and the ROS\_HOSTNAME set to the IP address of the Turtlebot3, ensuring efficient data exchange and control during the real-world experiments.

\subsection{Simultaneous Localization and Mapping}
\begin{figure}[htp]
\centering
   \subfloat[]{\label{fig:slam_with_objects}
      \includegraphics[width=.22\textwidth]{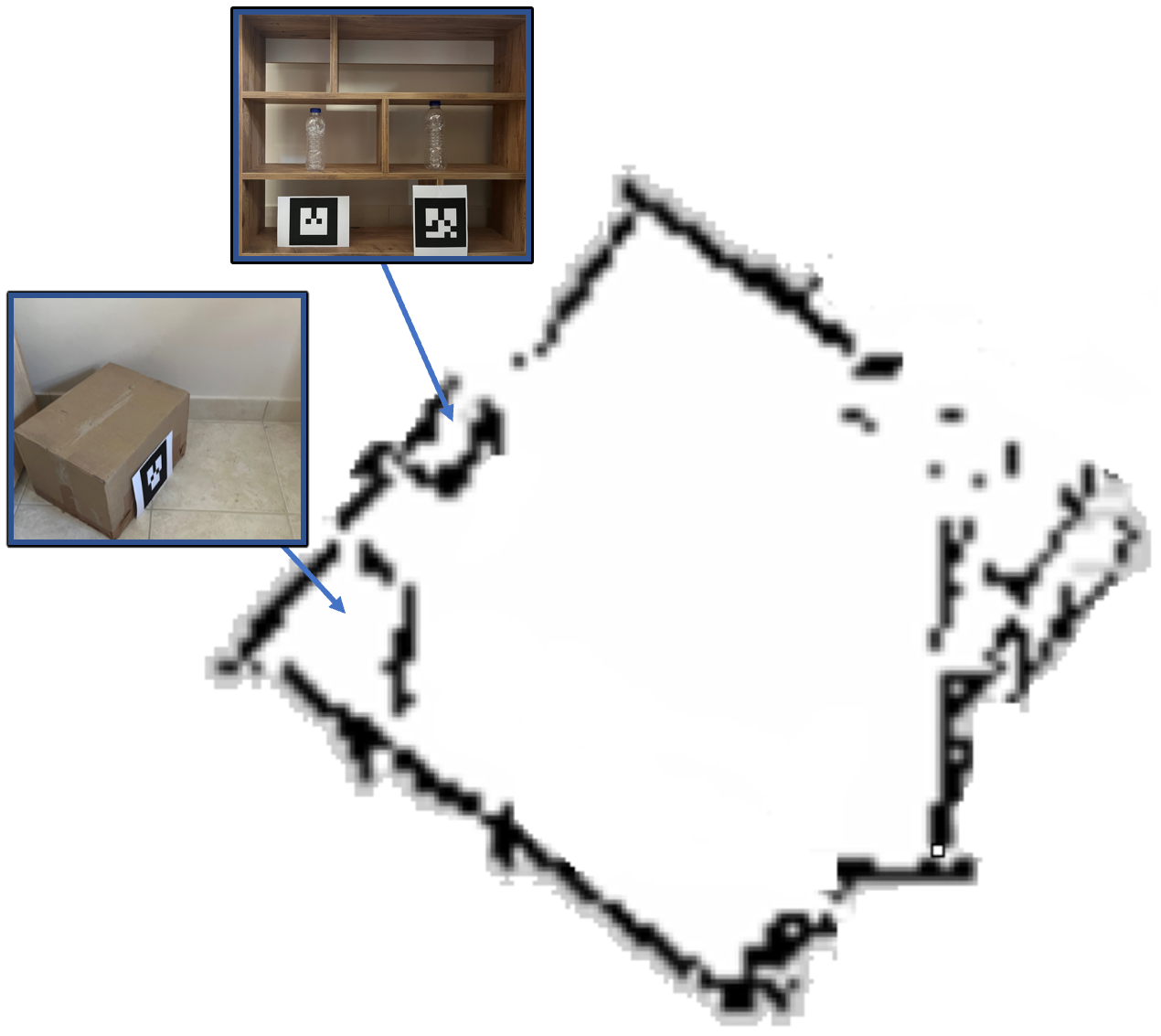}}
   \subfloat[]{\label{fig:ar_marker_detection_real}
      \includegraphics[width=.22\textwidth]{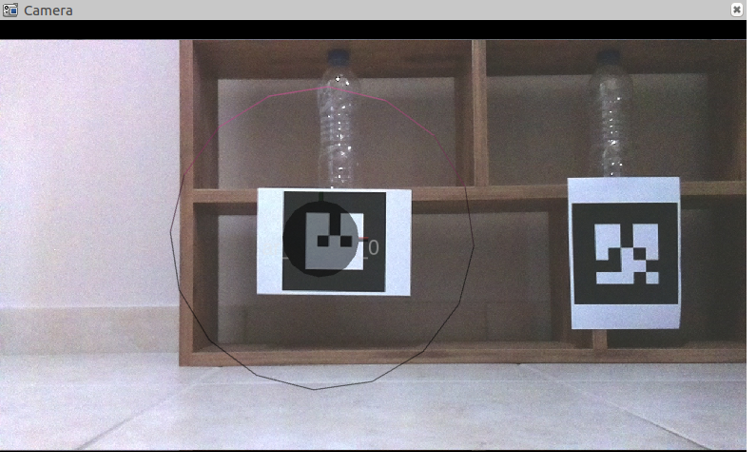}}
   \caption{Environment layout. In (a) the simultaneous localization and mapping of constructing a map of an unknown environment is presented while (b) provides a visual representation of the robot's camera during the AR marker detection.}\label{}
\end{figure}

To perform a real-world experiment, the robot must first navigate and explore an unknown environment with the objective of creating a map of its surroundings. In this paper, for constructing our own experimental environment, we utilize the well-established SLAM technique \cite{aulinas2008slam}. Upon deployment, the robot begins its exploration, relying on a combination of a 180-degree LDS-02 lidar sensor, a high-resolution camera, and wheel encoders for precise motion tracking. As the TurtleBot3 navigates through space, it continuously collects data from its sensors while scanning the surroundings. Leveraging real-time data, the robot intelligently links new observations with existing map features, refining its understanding of the environment's layout, and ultimately constructing a map of a room measuring 4 by 3 meters, as depicted in Fig. \ref{fig:slam_with_objects}.


\subsection{Sim-to-Real World Knowledge Transfer}
\begin{figure}[htp]
\centering
\includegraphics[width=0.95\columnwidth]{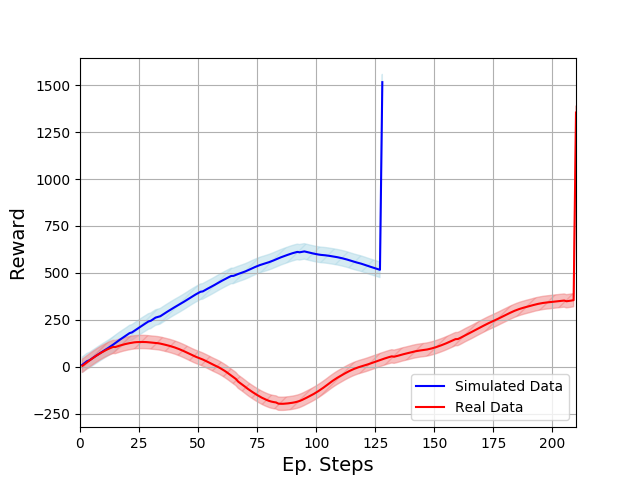}
\caption{Reinforcement learning performance comparison between simulation and real-world environments. The blue line depicts task performance in the simulation environment and the red line showcases the results after the sim-to-real best policy transfer.}
\label{fig:trasfer_learning}
\end{figure}

Having constructed our own custom map, our goal is to execute the home service scenario, as described in subsection \ref{ref:home_service_scenario}. As a first step, the scenario requires the robot navigation to a target location and more specifically to the kitchen area (`Go to the Kitchen'). By translating natural language to robotic actions, GPT-4 uses the \textit{get\_position} function to specify the target's location ($x, y$), as depicted in Fig. \ref{fig:prompt}. Once the location is specified, the RL-based navigation function described in section \ref{sec::navigation} is employed by transferring the knowledge learned by the agent (see fig. \ref{fig::rl_perfomance}) in the simulation environment (see fig. \ref{fig::gazebo}) to the real-world setup (see fig. \ref{fig:slam_with_objects}). 
Having defined the best-performing RL algorithm (PPO), we focus on the policy transfer from the simulation training, aiming to showcase the performance of the agent's policy in both simulated and real-world environments until successfully reaching the desired target location. It should be noted that the overall system characteristics (e.g., robot model, sensor noise, camera resolution, LiDAR's range, etc.) were maintained consistent in both simulated and real-world environments. The key distinction between these two lies in the location of objects, where the target location in the real world differs from that used during the agent's training in the simulation. Fig. \ref{fig:trasfer_learning} presents a comparative assessment of the agent's policy performance in the sim-to-real transfer learning process.

The observed discrepancy between the real-world and the simulation data, particularly the initial drop and subsequent increase, can be attributed to several factors inherent in the Sim-to-Real transfer process.

In real-world scenarios, environmental factors such as sensor noise, unmodeled dynamics, or slight variations in robot behavior can lead to deviations from the idealized conditions simulated during training. The initial drop in performance reflects the system's adjustment to these real-world conditions, where the learned policy encounters differences that it did not experience during simulation. As the system continues to operate, it gradually adapts, resulting in the observed increase in performance.

The final spike in the curve indicates a sudden adjustment or correction made by the robot as it converges on the learned policy. More specifically, this spike occurred when the robot encountered a previously unseen scenario related to the target location, which it managed effectively due to its learned policy, combined with an abrupt improvement in sensor readings or environmental conditions.



This behavior highlights the complexity and challenges of Sim-to-Real transfer, where perfect alignment between simulated and real-world performance is difficult to achieve. However, despite these variations, the overall trend demonstrates successful knowledge transfer, with the system adapting to and eventually improving over time, despite the discrepancies between the two worlds. 

\subsection{Target Alignment}
Upon reaching the target, the AR marker detection and pose estimation function is executed to validate the accurate alignment of the robot within the kitchen area and in proximity to the intended object. As depicted in Fig. \ref{fig:ar_marker_detection_real}, the robot effectively identifies and monitors the fiducial marker. Utilizing the camera's position and orientation, we obtain the desired pose in which the water bottle is positioned by revealing the fiducial marker's location.

\subsection{Pick-Grasp-Place}
\begin{figure}[htp]
\centering
   \subfloat[\centering Pre-pick arm position.]{\label{fig::wait_to_pick}
      \includegraphics[width=.147\textwidth]{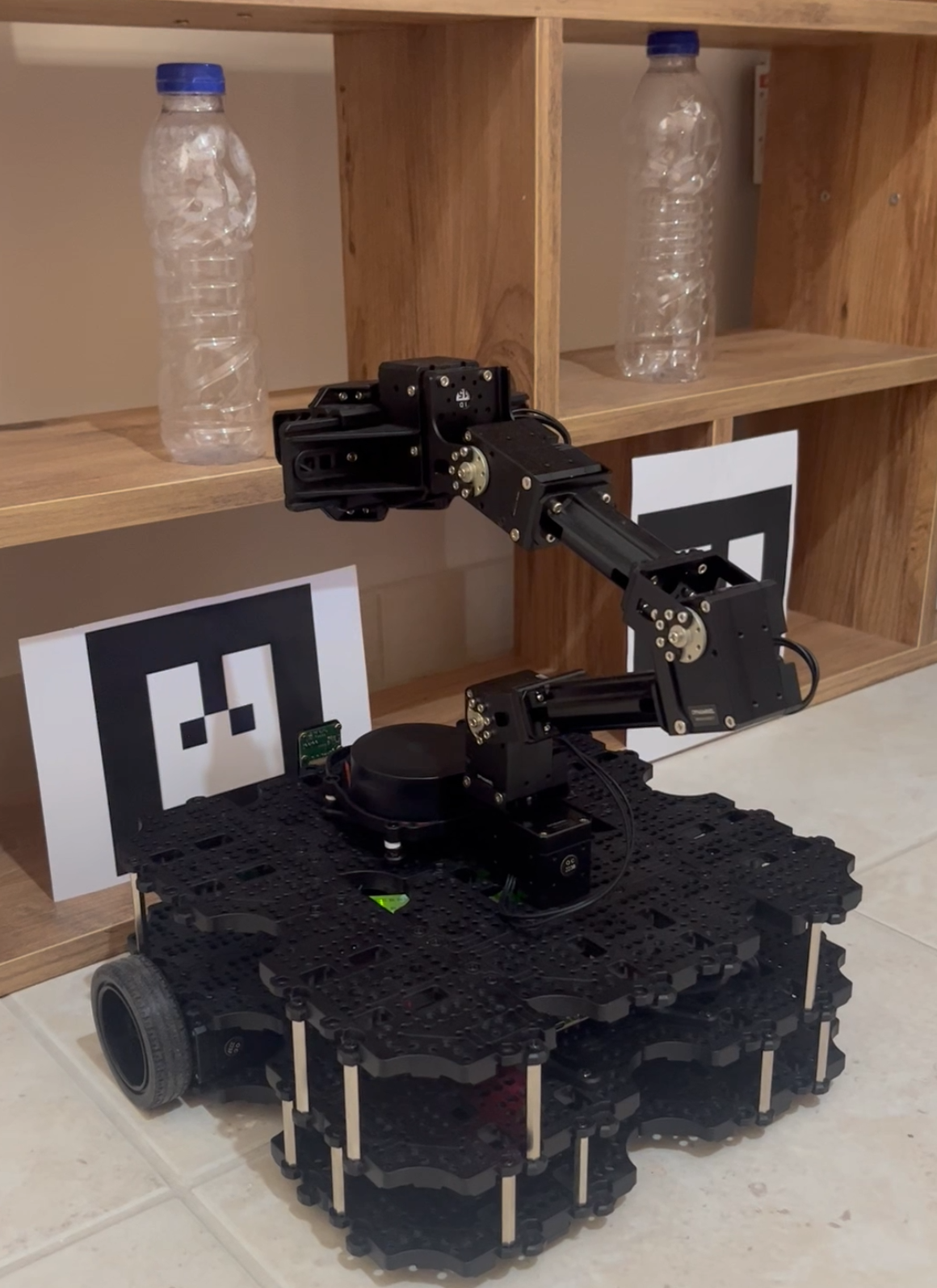}}
   \subfloat[\centering Pick arm \newline position.]{\label{fig::pick_arm_position}
      \includegraphics[width=.15\textwidth]{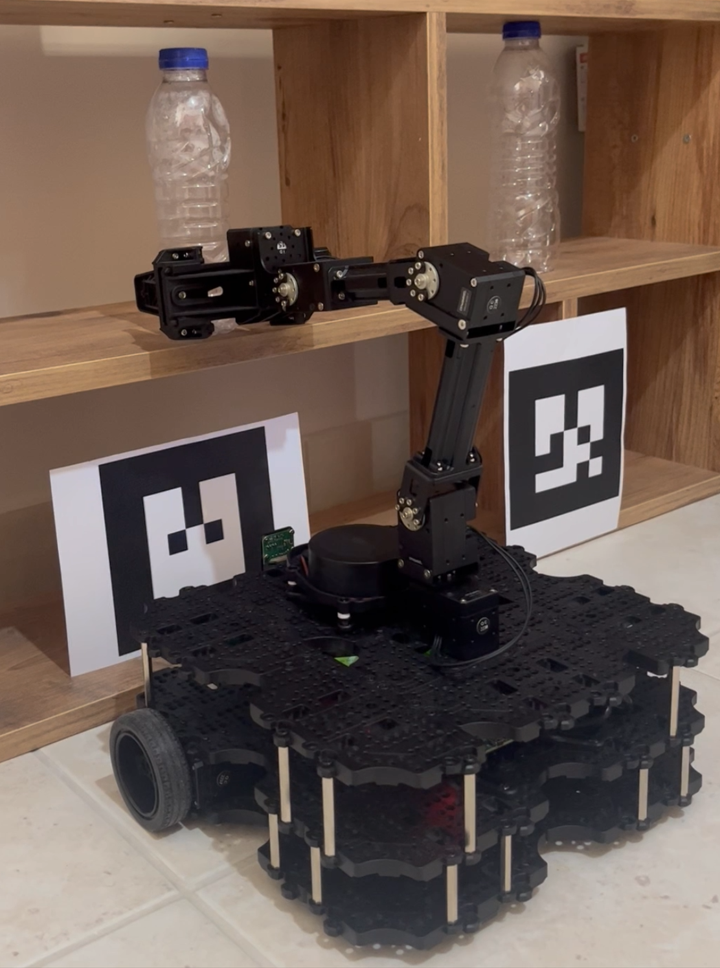}}
   \subfloat[\centering Post-pick arm position.]{\label{fig::Post-pick arm position}
      \includegraphics[width=.15\textwidth]{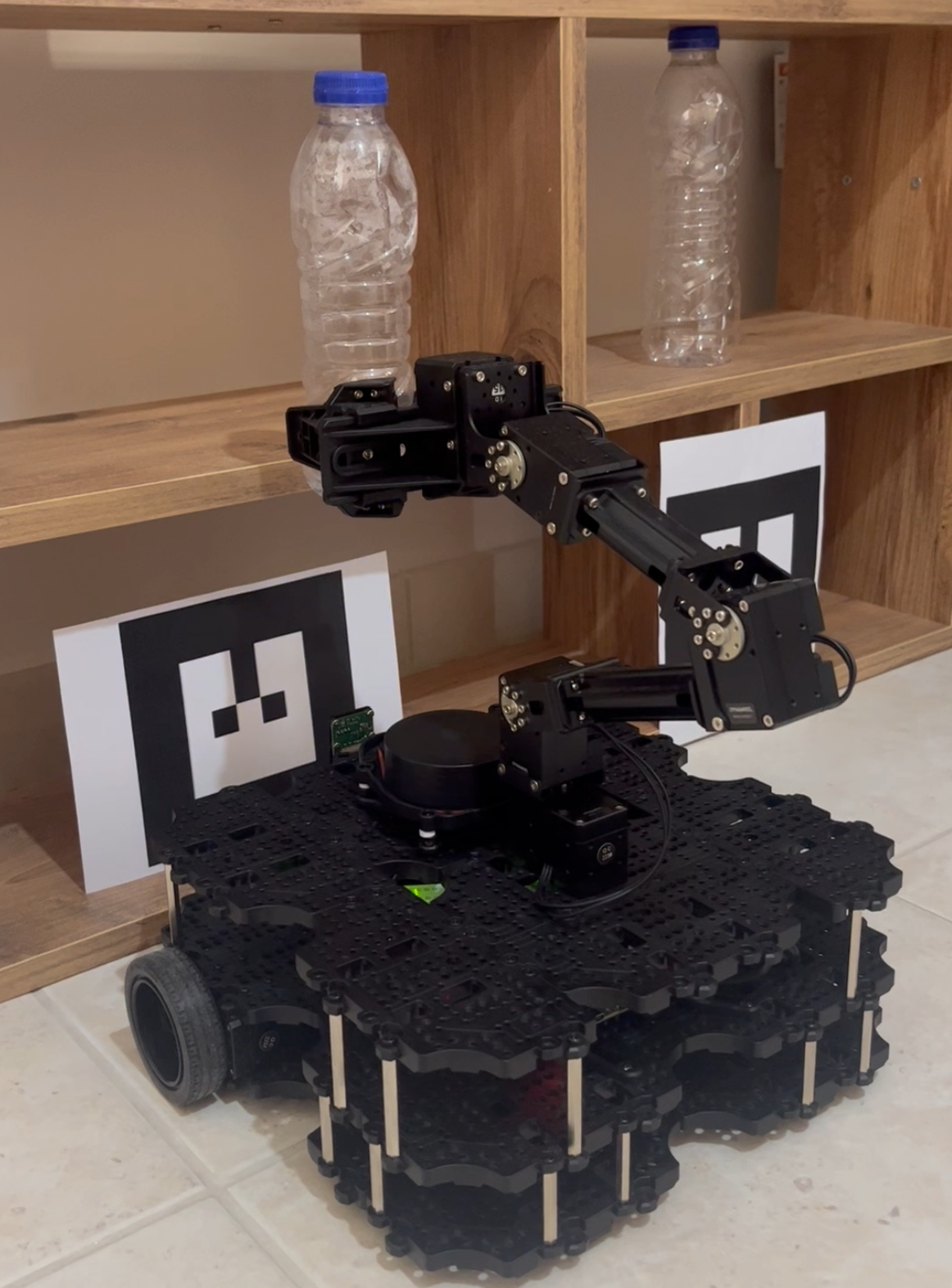}}

   \subfloat[\centering Pre-place arm position]{\label{fig::Pre-place arm position}
      \includegraphics[width=.146\textwidth]{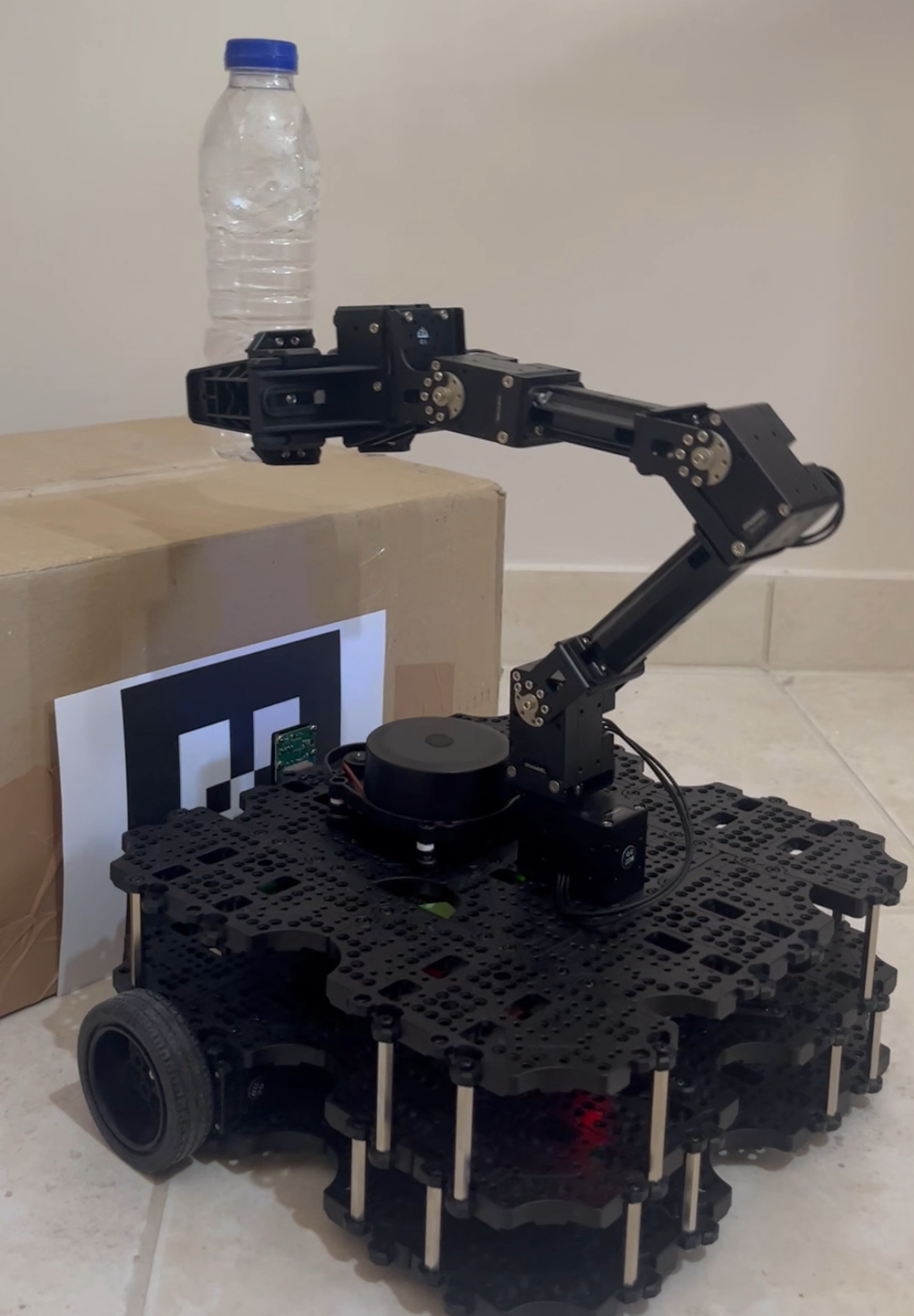}}
   \subfloat[\centering Place arm \newline position]{\label{fig::Place arm position}
      \includegraphics[width=.1505\textwidth]{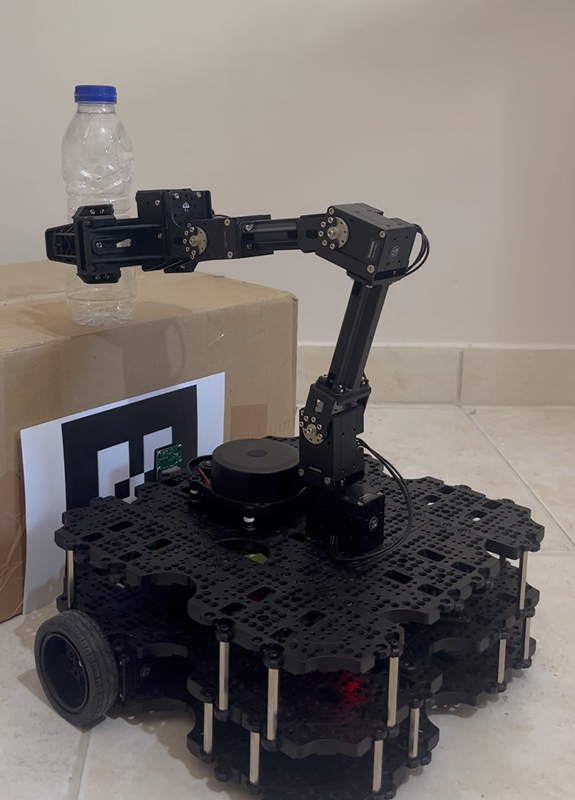}}
   \subfloat[\centering Post-place arm position]{\label{fig::Post-place arm position}
      \includegraphics[width=.150\textwidth]{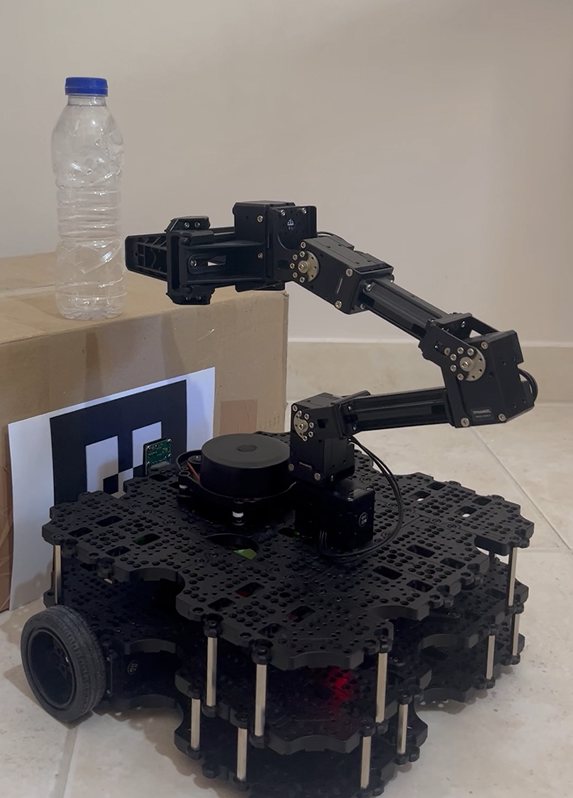}}

   \caption{Illustration of the sequential steps involved in a pick \& place manipulation task performed by the robotic arm.}\label{fig:manipulation_figures}
\end{figure}

Having established the object's location, GPT-4 is now capable of exerting control over the robot's actions to execute the pick-and-place manipulation task and deliver the bottle of water to the human. Fig. \ref{fig:manipulation_figures} depicts a step-by-step illustration of the pick-and-place manipulation task executed by the Turtlebot3. The process begins with the robot's arm in the ``Pre-pick arm position" (see fig. \ref{fig::wait_to_pick}), having the gripper closed and ready to initiate the object grasping operation. As the task commences, the robot moves into the ``Pick arm position" (see fig. \ref{fig::pick_arm_position}), where the gripper opens around the object securely in order to grasp it. Upon successful grasp (gripper with close fingers), the robot transitions into the ``Post-pick arm position" (see fig. \ref{fig::Post-pick arm position}), ensuring the object is firmly and safely held during transport. Subsequently, the robot navigates to its destination through sim-to-real knowledge transfer, as depicted in Fig. \ref{fig:trasfer_learning}, positioning the robot's arm in the ``Pre-place arm position" (see fig. \ref{fig::Pre-place arm position}). At the target location, the robot enters the ``Place arm position" (see fig. \ref{fig::Place arm position}), gently opening the gripper and placing the object in its designated spot (gripper with open fingers). Finally, the robot moves to the ``Post-place arm position" (see fig. \ref{fig::Post-place arm position}), ensuring that the robotic arm is ready for the next task having the gripper in the closed position. This visual representation highlights the key stages of pick and place manipulation tasks, exemplifying the precision and adaptability of the proposed robotic system in executing complex tasks over GPT-4 commands.

\subsubsection{Qualitative Analysis of Grasp Planning and Execution}

\begin{figure*}[htp]
\centering
   \subfloat[Gripper with fingers in an opened position.]{\label{fig::gripper_closed}
      \includegraphics[width=.45\textwidth]{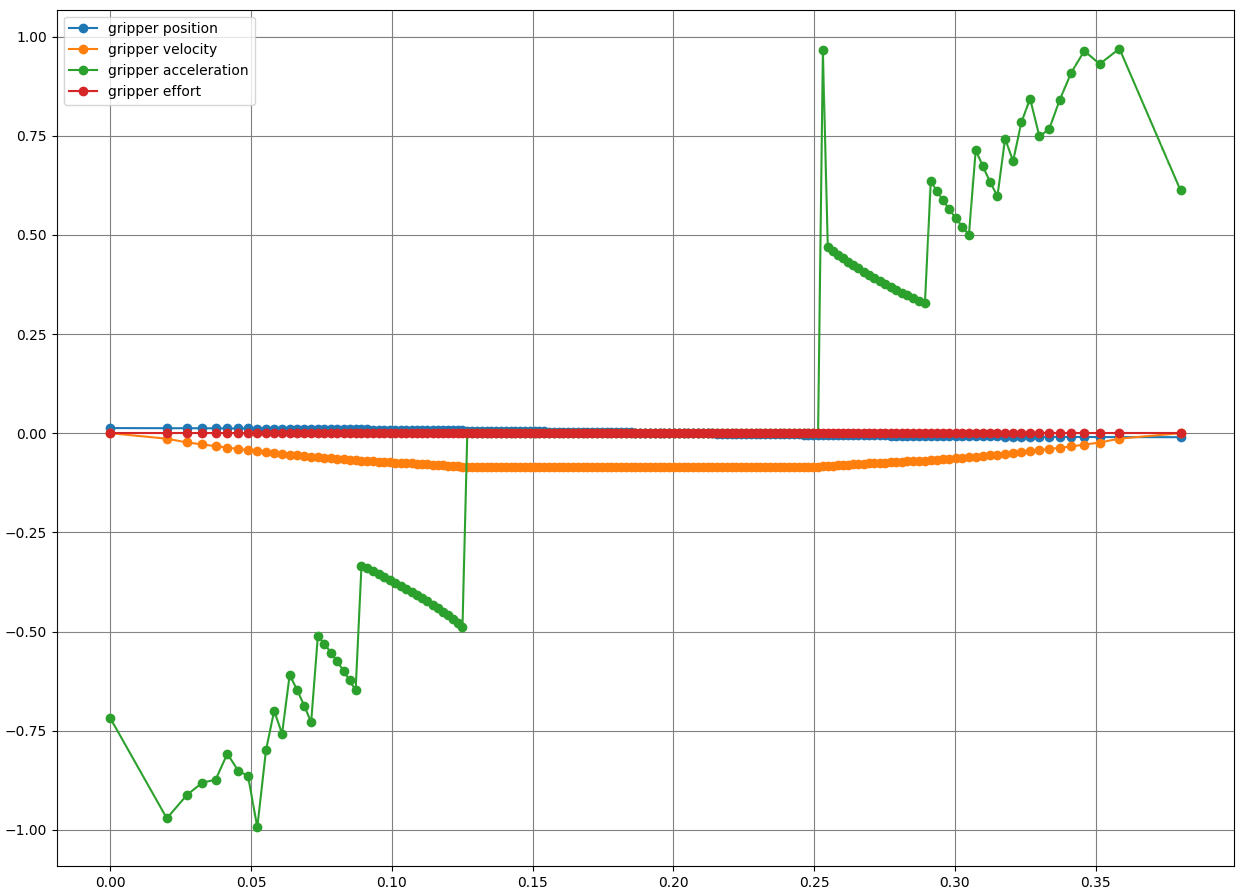}}
   \subfloat[Gripper with fingers in an opened position]{\label{fig::gripper_opened}
      \includegraphics[width=.45\textwidth]{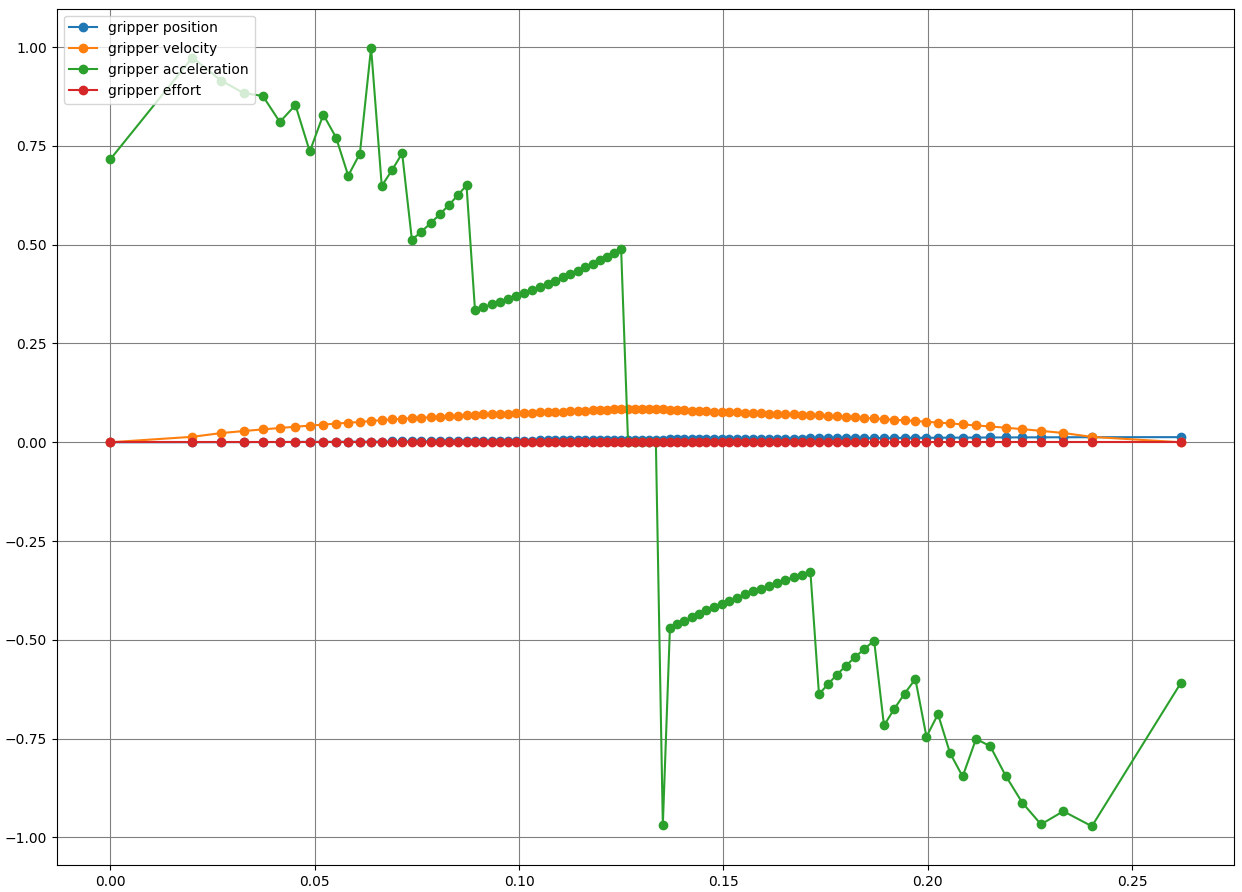}}
   \caption{\centering Trajectories for the robot's gripper joints during pick \& place. \newline
    \centering
    ($x \rightarrow s$; $y \rightarrow m$, $\frac{m}{s}$, $\frac{m}{s^2}$, $N$)}\label{fig:rqt_figures}
\end{figure*}

To qualitatively analyze the robot's performance during the pick and place tasks, trajectory plots of the gripper's joints were generated using the rqt tool \cite{rqt}. These plots offer a clear visualization of the robot's gripper joint movements and the accuracy of the executed trajectories. Fig. \ref{fig:rqt_figures} displays two graphs representing real-time data of the gripper's joint trajectories during both the picking and placing actions. X-axis represents time in seconds ($s$), scaled to the duration of the robot's actions, while the y-axis displays the corresponding units of measurement, including meters ($m$) for position, meters per second ($\frac{m}{s}$) for velocity, meters per second squared ($\frac{m}{s^2}$) for acceleration, and newtons ($N$) for effort.


To have a deeper understanding of the graphs, it is necessary to keep in mind that the reference direction of the specified quantities originates from the center of the gripper (where the fingers are located when it's closed) and extends outward (to the position of the fingers when the gripper is open).

While GPT-4 generates a plan based on user input and issues a command to control the gripper (for either picking or placing), the effort required from the gripper remains consistently null throughout the entire duration of the robot's actions. The most notable variation in both figures is that of the acceleration. During the picking phase (when the gripper is closed - see fig. \ref{fig::gripper_closed}), the acceleration rapidly increases from zero to a peak modulus of nearly 1 $\frac{m}{s^2}$. This produces a coherent sign velocity, causing the joint position value to decrease as the joint approaches the center of the gripper. The acceleration subsequently returns to zero, resulting in the finger moving at a constant speed until it nearly reaches the correct position to lock the object. Approximately a tenth of a second before reaching the expected position, the acceleration increases with the opposite sign of the velocity, leading to deceleration and ultimately halting the finger. Once the desired position is reached, the action is completed, and the movement stops without requiring the acceleration to return to zero. On the other hand, during the placing phase (when the gripper is opened - see fig. \ref{fig::gripper_opened}), the same process occurs, but in reverse. The fingers move in opposite directions due to the opposite directions of velocity and acceleration.

A video demonstration showcasing the real-world experiments within our custom environment involving the robotic manipulation tasks over GPT-4 is attached to the following link\footnote{\url{https://youtu.be/WgfPbgZK8ic}}.

\section{Conclusions} 
\label{sec:conclusions}
This paper introduces an innovative advancement in robotics, uniting state-of-the-art AI technology with modular robotic functionalities within a ROS-based framework. The key innovation lies in the seamless translation of natural language into precise robotic actions while marking the pioneering of a custom robot library suite with GPT-4 into a ROS-based robotic system. Particularly highlighted through comprehensive experiments both in simulated and in real-world environments, RobotIQ yielded remarkable outcomes, as evidenced by accurate navigation achieved through the best learning-based policy transfer from simulation to real-world, object localization, autonomous manipulation, and successful pick-and-place actions.

Empowering mobile robots with human-level planning utilizing LLM models sets a new standard for effective human-robot interaction. The RobotIQ framework, designed with a modular architecture within the ROS ecosystem, is inherently adaptable to various robotic platforms beyond the Turtlebot3 Waffle Pi used in our experiments. This adaptability comes from using standardized ROS messages and interfaces, which facilitate the integration of different sensors, actuators, and computational units. However, transferring the system to other real robots presents several challenges. These include hardware compatibility issues, such as differences in sensor specifications, actuator dynamics, and computational power, which may necessitate modifications to the control algorithms and recalibration of sensor data processing. Additionally, the physical configuration and operational constraints of different robots might require customized adjustments to the navigation and manipulation modules to ensure robust performance in diverse environments. Addressing these challenges involves extensive testing and fine-tuning, but the open-source nature of RobotIQ allows for community-driven enhancements and support for a broader range of robotic platforms.

\section*{Acknowledgements}
This research has been co-financed by the European Union and Greek national funds through the Operational Program Competitiveness, Entrepreneurship and Innovation, under the call RESEARCH-CREATE-INNOVATE (T2EDK-02743).

\section*{Statements and Declarations}
\textbf{Ethical Approval} 
Not applicable, as our paper does not encompass studies involving humans or animals.\\
\\
\textbf{Funding} 
This research has been co-financed by the European Union and Greek national funds through the Operational Program Competitiveness, Entrepreneurship and Innovation, under the call RESEARCH-CREATE-INNOVATE (T2EDK-02743).  \\
\\
\textbf{Availability of data and materials} 
For comprehensive access to our research materials, including the algorithm, evaluation mechanisms, and results, please refer to the repository at  https://github.com/emmarapt/RobotIQ.git, where all relevant data is available. This open-source repository serves as a central hub for accessing and exploring the complete set of resources associated with our study.\\
\\
\textbf{Conflict of Interests}
The authors declare that they have no known competing financial interests or personal relationships that could have appeared to influence the work reported in this paper.


\bibliography{references.bib}


\begin{thebibliography}{48}
\ifx \bisbn   \undefined \def \bisbn  #1{ISBN #1}\fi
\ifx \binits  \undefined \def \binits#1{#1}\fi
\ifx \bauthor  \undefined \def \bauthor#1{#1}\fi
\ifx \batitle  \undefined \def \batitle#1{#1}\fi
\ifx \bjtitle  \undefined \def \bjtitle#1{#1}\fi
\ifx \bvolume  \undefined \def \bvolume#1{\textbf{#1}}\fi
\ifx \byear  \undefined \def \byear#1{#1}\fi
\ifx \bissue  \undefined \def \bissue#1{#1}\fi
\ifx \bfpage  \undefined \def \bfpage#1{#1}\fi
\ifx \blpage  \undefined \def \blpage #1{#1}\fi
\ifx \burl  \undefined \def \burl#1{\textsf{#1}}\fi
\ifx \doiurl  \undefined \def \doiurl#1{\url{https://doi.org/#1}}\fi
\ifx \betal  \undefined \def \betal{\textit{et al.}}\fi
\ifx \binstitute  \undefined \def \binstitute#1{#1}\fi
\ifx \binstitutionaled  \undefined \def \binstitutionaled#1{#1}\fi
\ifx \bctitle  \undefined \def \bctitle#1{#1}\fi
\ifx \beditor  \undefined \def \beditor#1{#1}\fi
\ifx \bpublisher  \undefined \def \bpublisher#1{#1}\fi
\ifx \bbtitle  \undefined \def \bbtitle#1{#1}\fi
\ifx \bedition  \undefined \def \bedition#1{#1}\fi
\ifx \bseriesno  \undefined \def \bseriesno#1{#1}\fi
\ifx \blocation  \undefined \def \blocation#1{#1}\fi
\ifx \bsertitle  \undefined \def \bsertitle#1{#1}\fi
\ifx \bsnm \undefined \def \bsnm#1{#1}\fi
\ifx \bsuffix \undefined \def \bsuffix#1{#1}\fi
\ifx \bparticle \undefined \def \bparticle#1{#1}\fi
\ifx \barticle \undefined \def \barticle#1{#1}\fi
\bibcommenthead
\ifx \bconfdate \undefined \def \bconfdate #1{#1}\fi
\ifx \botherref \undefined \def \botherref #1{#1}\fi
\ifx \url \undefined \def \url#1{\textsf{#1}}\fi
\ifx \bchapter \undefined \def \bchapter#1{#1}\fi
\ifx \bbook \undefined \def \bbook#1{#1}\fi
\ifx \bcomment \undefined \def \bcomment#1{#1}\fi
\ifx \oauthor \undefined \def \oauthor#1{#1}\fi
\ifx \citeauthoryear \undefined \def \citeauthoryear#1{#1}\fi
\ifx \endbibitem  \undefined \def \endbibitem {}\fi
\ifx \bconflocation  \undefined \def \bconflocation#1{#1}\fi
\ifx \arxivurl  \undefined \def \arxivurl#1{\textsf{#1}}\fi
\csname PreBibitemsHook\endcsname

\bibitem[\protect\citeauthoryear{Liu et~al.}{2017}]{liu2017intelligence}
\begin{barticle}
\bauthor{\bsnm{Liu}, \binits{F.}},
\bauthor{\bsnm{Shi}, \binits{Y.}},
\bauthor{\bsnm{Liu}, \binits{Y.}}:
\batitle{Intelligence quotient and intelligence grade of artificial intelligence}.
\bjtitle{Annals of Data Science}
\bvolume{4},
\bfpage{179}--\blpage{191}
(\byear{2017})
\end{barticle}
\endbibitem

\bibitem[\protect\citeauthoryear{Vaswani et~al.}{2017}]{vaswani2017attention}
\begin{botherref}
\oauthor{\bsnm{Vaswani}, \binits{A.}},
\oauthor{\bsnm{Shazeer}, \binits{N.}},
\oauthor{\bsnm{Parmar}, \binits{N.}},
\oauthor{\bsnm{Uszkoreit}, \binits{J.}},
\oauthor{\bsnm{Jones}, \binits{L.}},
\oauthor{\bsnm{Gomez}, \binits{A.N.}},
\oauthor{\bsnm{Kaiser}, \binits{{\L}.}},
\oauthor{\bsnm{Polosukhin}, \binits{I.}}:
Attention is all you need.
Advances in neural information processing systems
\textbf{30}
(2017)
\end{botherref}
\endbibitem

\bibitem[\protect\citeauthoryear{OpenAI}{2023}]{ChatGPT}
\begin{botherref}
\oauthor{\bsnm{OpenAI}}:
ChatGPT.
\url{https://openai.com/blog/chatgpt}.
[Online; accessed 26 July 2023]
(2023)
\end{botherref}
\endbibitem

\bibitem[\protect\citeauthoryear{Liu et~al.}{2023}]{liu2023pre}
\begin{barticle}
\bauthor{\bsnm{Liu}, \binits{P.}},
\bauthor{\bsnm{Yuan}, \binits{W.}},
\bauthor{\bsnm{Fu}, \binits{J.}},
\bauthor{\bsnm{Jiang}, \binits{Z.}},
\bauthor{\bsnm{Hayashi}, \binits{H.}},
\bauthor{\bsnm{Neubig}, \binits{G.}}:
\batitle{Pre-train, prompt, and predict: A systematic survey of prompting methods in natural language processing}.
\bjtitle{ACM Computing Surveys}
\bvolume{55}(\bissue{9}),
\bfpage{1}--\blpage{35}
(\byear{2023})
\end{barticle}
\endbibitem

\bibitem[\protect\citeauthoryear{Huang et~al.}{2023}]{huang2023voxposer}
\begin{bchapter}
\bauthor{\bsnm{Huang}, \binits{W.}},
\bauthor{\bsnm{Wang}, \binits{C.}},
\bauthor{\bsnm{Zhang}, \binits{R.}},
\bauthor{\bsnm{Li}, \binits{Y.}},
\bauthor{\bsnm{Wu}, \binits{J.}},
\bauthor{\bsnm{Fei-Fei}, \binits{L.}}:
\bctitle{Voxposer: Composable 3d value maps for robotic manipulation with language models}.
In: \bbtitle{Conference on Robot Learning},
pp. \bfpage{540}--\blpage{562}
(\byear{2023}).
\bcomment{PMLR}
\end{bchapter}
\endbibitem

\bibitem[\protect\citeauthoryear{Liang et~al.}{2023}]{liang2023code}
\begin{bchapter}
\bauthor{\bsnm{Liang}, \binits{J.}},
\bauthor{\bsnm{Huang}, \binits{W.}},
\bauthor{\bsnm{Xia}, \binits{F.}},
\bauthor{\bsnm{Xu}, \binits{P.}},
\bauthor{\bsnm{Hausman}, \binits{K.}},
\bauthor{\bsnm{Ichter}, \binits{B.}},
\bauthor{\bsnm{Florence}, \binits{P.}},
\bauthor{\bsnm{Zeng}, \binits{A.}}:
\bctitle{Code as policies: Language model programs for embodied control}.
In: \bbtitle{2023 IEEE International Conference on Robotics and Automation (ICRA)},
pp. \bfpage{9493}--\blpage{9500}
(\byear{2023}).
\bcomment{IEEE}
\end{bchapter}
\endbibitem

\bibitem[\protect\citeauthoryear{Huang et~al.}{2023}]{huang2023grounded}
\begin{botherref}
\oauthor{\bsnm{Huang}, \binits{W.}},
\oauthor{\bsnm{Xia}, \binits{F.}},
\oauthor{\bsnm{Shah}, \binits{D.}},
\oauthor{\bsnm{Driess}, \binits{D.}},
\oauthor{\bsnm{Zeng}, \binits{A.}},
\oauthor{\bsnm{Lu}, \binits{Y.}},
\oauthor{\bsnm{Florence}, \binits{P.}},
\oauthor{\bsnm{Mordatch}, \binits{I.}},
\oauthor{\bsnm{Levine}, \binits{S.}},
\oauthor{\bsnm{Hausman}, \binits{K.}}, et al.:
Grounded decoding: Guiding text generation with grounded models for robot control.
arXiv preprint arXiv:2303.00855
(2023)
\end{botherref}
\endbibitem

\bibitem[\protect\citeauthoryear{Huang et~al.}{2021}]{huang2021generalization}
\begin{botherref}
\oauthor{\bsnm{Huang}, \binits{W.}},
\oauthor{\bsnm{Mordatch}, \binits{I.}},
\oauthor{\bsnm{Abbeel}, \binits{P.}},
\oauthor{\bsnm{Pathak}, \binits{D.}}:
Generalization in dexterous manipulation via geometry-aware multi-task learning.
arXiv preprint arXiv:2111.03062
(2021)
\end{botherref}
\endbibitem

\bibitem[\protect\citeauthoryear{Zhou et~al.}{2023}]{zhou2023isr}
\begin{botherref}
\oauthor{\bsnm{Zhou}, \binits{Z.}},
\oauthor{\bsnm{Song}, \binits{J.}},
\oauthor{\bsnm{Yao}, \binits{K.}},
\oauthor{\bsnm{Shu}, \binits{Z.}},
\oauthor{\bsnm{Ma}, \binits{L.}}:
Isr-llm: Iterative self-refined large language model for long-horizon sequential task planning.
arXiv preprint arXiv:2308.13724
(2023)
\end{botherref}
\endbibitem

\bibitem[\protect\citeauthoryear{Tellex et~al.}{2020}]{tellex2020robots}
\begin{barticle}
\bauthor{\bsnm{Tellex}, \binits{S.}},
\bauthor{\bsnm{Gopalan}, \binits{N.}},
\bauthor{\bsnm{Kress-Gazit}, \binits{H.}},
\bauthor{\bsnm{Matuszek}, \binits{C.}}:
\batitle{Robots that use language}.
\bjtitle{Annual Review of Control, Robotics, and Autonomous Systems}
\bvolume{3}(\bissue{1}),
\bfpage{25}--\blpage{55}
(\byear{2020})
\end{barticle}
\endbibitem

\bibitem[\protect\citeauthoryear{Huang et~al.}{2022}]{huang2022language}
\begin{bchapter}
\bauthor{\bsnm{Huang}, \binits{W.}},
\bauthor{\bsnm{Abbeel}, \binits{P.}},
\bauthor{\bsnm{Pathak}, \binits{D.}},
\bauthor{\bsnm{Mordatch}, \binits{I.}}:
\bctitle{Language models as zero-shot planners: Extracting actionable knowledge for embodied agents}.
In: \bbtitle{International Conference on Machine Learning},
pp. \bfpage{9118}--\blpage{9147}
(\byear{2022}).
\bcomment{PMLR}
\end{bchapter}
\endbibitem

\bibitem[\protect\citeauthoryear{Hao et~al.}{2023}]{hao2023reasoning}
\begin{botherref}
\oauthor{\bsnm{Hao}, \binits{S.}},
\oauthor{\bsnm{Gu}, \binits{Y.}},
\oauthor{\bsnm{Ma}, \binits{H.}},
\oauthor{\bsnm{Hong}, \binits{J.J.}},
\oauthor{\bsnm{Wang}, \binits{Z.}},
\oauthor{\bsnm{Wang}, \binits{D.Z.}},
\oauthor{\bsnm{Hu}, \binits{Z.}}:
Reasoning with language model is planning with world model.
arXiv preprint arXiv:2305.14992
(2023)
\end{botherref}
\endbibitem

\bibitem[\protect\citeauthoryear{Jin et~al.}{2023}]{jin2023alphablock}
\begin{botherref}
\oauthor{\bsnm{Jin}, \binits{C.}},
\oauthor{\bsnm{Tan}, \binits{W.}},
\oauthor{\bsnm{Yang}, \binits{J.}},
\oauthor{\bsnm{Liu}, \binits{B.}},
\oauthor{\bsnm{Song}, \binits{R.}},
\oauthor{\bsnm{Wang}, \binits{L.}},
\oauthor{\bsnm{Fu}, \binits{J.}}:
Alphablock: Embodied finetuning for vision-language reasoning in robot manipulation.
arXiv preprint arXiv:2305.18898
(2023)
\end{botherref}
\endbibitem

\bibitem[\protect\citeauthoryear{Singh et~al.}{2023}]{singh2023progprompt}
\begin{bchapter}
\bauthor{\bsnm{Singh}, \binits{I.}},
\bauthor{\bsnm{Blukis}, \binits{V.}},
\bauthor{\bsnm{Mousavian}, \binits{A.}},
\bauthor{\bsnm{Goyal}, \binits{A.}},
\bauthor{\bsnm{Xu}, \binits{D.}},
\bauthor{\bsnm{Tremblay}, \binits{J.}},
\bauthor{\bsnm{Fox}, \binits{D.}},
\bauthor{\bsnm{Thomason}, \binits{J.}},
\bauthor{\bsnm{Garg}, \binits{A.}}:
\bctitle{Progprompt: Generating situated robot task plans using large language models}.
In: \bbtitle{2023 IEEE International Conference on Robotics and Automation (ICRA)},
pp. \bfpage{11523}--\blpage{11530}
(\byear{2023}).
\bcomment{IEEE}
\end{bchapter}
\endbibitem

\bibitem[\protect\citeauthoryear{Huang et~al.}{2022}]{huang2022inner}
\begin{botherref}
\oauthor{\bsnm{Huang}, \binits{W.}},
\oauthor{\bsnm{Xia}, \binits{F.}},
\oauthor{\bsnm{Xiao}, \binits{T.}},
\oauthor{\bsnm{Chan}, \binits{H.}},
\oauthor{\bsnm{Liang}, \binits{J.}},
\oauthor{\bsnm{Florence}, \binits{P.}},
\oauthor{\bsnm{Zeng}, \binits{A.}},
\oauthor{\bsnm{Tompson}, \binits{J.}},
\oauthor{\bsnm{Mordatch}, \binits{I.}},
\oauthor{\bsnm{Chebotar}, \binits{Y.}}, et al.:
Inner monologue: Embodied reasoning through planning with language models.
arXiv preprint arXiv:2207.05608
(2022)
\end{botherref}
\endbibitem

\bibitem[\protect\citeauthoryear{Song et~al.}{2023}]{song2023llm}
\begin{bchapter}
\bauthor{\bsnm{Song}, \binits{C.H.}},
\bauthor{\bsnm{Wu}, \binits{J.}},
\bauthor{\bsnm{Washington}, \binits{C.}},
\bauthor{\bsnm{Sadler}, \binits{B.M.}},
\bauthor{\bsnm{Chao}, \binits{W.-L.}},
\bauthor{\bsnm{Su}, \binits{Y.}}:
\bctitle{Llm-planner: Few-shot grounded planning for embodied agents with large language models}.
In: \bbtitle{Proceedings of the IEEE/CVF International Conference on Computer Vision},
pp. \bfpage{2998}--\blpage{3009}
(\byear{2023})
\end{bchapter}
\endbibitem

\bibitem[\protect\citeauthoryear{Ding et~al.}{}]{dingtask}
\begin{botherref}
\oauthor{\bsnm{Ding}, \binits{Y.}},
\oauthor{\bsnm{Zhang}, \binits{X.}},
\oauthor{\bsnm{Paxton}, \binits{C.}},
\oauthor{\bsnm{Zhang}, \binits{S.}}:
Task and motion planning with large language models for object rearrangement. in 2023 ieee.
In: RSJ International Conference on Intelligent Robots and Systems (IROS),
pp. 2086--2092
\end{botherref}
\endbibitem

\bibitem[\protect\citeauthoryear{Liu et~al.}{2024}]{liu2024delta}
\begin{botherref}
\oauthor{\bsnm{Liu}, \binits{Y.}},
\oauthor{\bsnm{Palmieri}, \binits{L.}},
\oauthor{\bsnm{Koch}, \binits{S.}},
\oauthor{\bsnm{Georgievski}, \binits{I.}},
\oauthor{\bsnm{Aiello}, \binits{M.}}:
Delta: Decomposed efficient long-term robot task planning using large language models.
arXiv preprint arXiv:2404.03275
(2024)
\end{botherref}
\endbibitem

\bibitem[\protect\citeauthoryear{Xu et~al.}{2018}]{xu2018neural}
\begin{bchapter}
\bauthor{\bsnm{Xu}, \binits{D.}},
\bauthor{\bsnm{Nair}, \binits{S.}},
\bauthor{\bsnm{Zhu}, \binits{Y.}},
\bauthor{\bsnm{Gao}, \binits{J.}},
\bauthor{\bsnm{Garg}, \binits{A.}},
\bauthor{\bsnm{Fei-Fei}, \binits{L.}},
\bauthor{\bsnm{Savarese}, \binits{S.}}:
\bctitle{Neural task programming: Learning to generalize across hierarchical tasks}.
In: \bbtitle{2018 IEEE International Conference on Robotics and Automation (ICRA)},
pp. \bfpage{3795}--\blpage{3802}
(\byear{2018}).
\bcomment{IEEE}
\end{bchapter}
\endbibitem

\bibitem[\protect\citeauthoryear{Srinivas et~al.}{2018}]{srinivas2018universal}
\begin{bchapter}
\bauthor{\bsnm{Srinivas}, \binits{A.}},
\bauthor{\bsnm{Jabri}, \binits{A.}},
\bauthor{\bsnm{Abbeel}, \binits{P.}},
\bauthor{\bsnm{Levine}, \binits{S.}},
\bauthor{\bsnm{Finn}, \binits{C.}}:
\bctitle{Universal planning networks: Learning generalizable representations for visuomotor control}.
In: \bbtitle{International Conference on Machine Learning},
pp. \bfpage{4732}--\blpage{4741}
(\byear{2018}).
\bcomment{PMLR}
\end{bchapter}
\endbibitem

\bibitem[\protect\citeauthoryear{Akakzia et~al.}{}]{akakziagrounding}
\begin{botherref}
\oauthor{\bsnm{Akakzia}, \binits{A.}},
\oauthor{\bsnm{Colas}, \binits{C.}},
\oauthor{\bsnm{Oudeyer}, \binits{P.-Y.}},
\oauthor{\bsnm{CHETOUANI}, \binits{M.}},
\oauthor{\bsnm{Sigaud}, \binits{O.}}:
Grounding language to autonomously-acquired skills via goal generation.
In: International Conference on Learning Representations
\end{botherref}
\endbibitem

\bibitem[\protect\citeauthoryear{Ha et~al.}{2023}]{ha2023scaling}
\begin{bchapter}
\bauthor{\bsnm{Ha}, \binits{H.}},
\bauthor{\bsnm{Florence}, \binits{P.}},
\bauthor{\bsnm{Song}, \binits{S.}}:
\bctitle{Scaling up and distilling down: Language-guided robot skill acquisition}.
In: \bbtitle{Conference on Robot Learning},
pp. \bfpage{3766}--\blpage{3777}
(\byear{2023}).
\bcomment{PMLR}
\end{bchapter}
\endbibitem

\bibitem[\protect\citeauthoryear{Rajvanshi et~al.}{2024}]{rajvanshi2024saynav}
\begin{bchapter}
\bauthor{\bsnm{Rajvanshi}, \binits{A.}},
\bauthor{\bsnm{Sikka}, \binits{K.}},
\bauthor{\bsnm{Lin}, \binits{X.}},
\bauthor{\bsnm{Lee}, \binits{B.}},
\bauthor{\bsnm{Chiu}, \binits{H.-P.}},
\bauthor{\bsnm{Velasquez}, \binits{A.}}:
\bctitle{Saynav: Grounding large language models for dynamic planning to navigation in new environments}.
In: \bbtitle{Proceedings of the International Conference on Automated Planning and Scheduling},
vol. \bseriesno{34},
pp. \bfpage{464}--\blpage{474}
(\byear{2024})
\end{bchapter}
\endbibitem

\bibitem[\protect\citeauthoryear{Dorbala et~al.}{2023}]{dorbala2023can}
\begin{botherref}
\oauthor{\bsnm{Dorbala}, \binits{V.S.}},
\oauthor{\bsnm{Mullen~Jr}, \binits{J.F.}},
\oauthor{\bsnm{Manocha}, \binits{D.}}:
Can an embodied agent find your “cat-shaped mug”? llm-based zero-shot object navigation.
IEEE Robotics and Automation Letters
(2023)
\end{botherref}
\endbibitem

\bibitem[\protect\citeauthoryear{Huang et~al.}{2023}]{huang2023visual}
\begin{bchapter}
\bauthor{\bsnm{Huang}, \binits{C.}},
\bauthor{\bsnm{Mees}, \binits{O.}},
\bauthor{\bsnm{Zeng}, \binits{A.}},
\bauthor{\bsnm{Burgard}, \binits{W.}}:
\bctitle{Visual language maps for robot navigation}.
In: \bbtitle{2023 IEEE International Conference on Robotics and Automation (ICRA)},
pp. \bfpage{10608}--\blpage{10615}
(\byear{2023}).
\bcomment{IEEE}
\end{bchapter}
\endbibitem

\bibitem[\protect\citeauthoryear{Vemprala et~al.}{2023}]{vemprala2023chatgpt}
\begin{barticle}
\bauthor{\bsnm{Vemprala}, \binits{S.}},
\bauthor{\bsnm{Bonatti}, \binits{R.}},
\bauthor{\bsnm{Bucker}, \binits{A.}},
\bauthor{\bsnm{Kapoor}, \binits{A.}}:
\batitle{Chatgpt for robotics: Design principles and model abilities}.
\bjtitle{Microsoft Auton. Syst. Robot. Res}
\bvolume{2},
\bfpage{20}
(\byear{2023})
\end{barticle}
\endbibitem

\bibitem[\protect\citeauthoryear{Chen et~al.}{2021}]{chen2021decision}
\begin{barticle}
\bauthor{\bsnm{Chen}, \binits{L.}},
\bauthor{\bsnm{Lu}, \binits{K.}},
\bauthor{\bsnm{Rajeswaran}, \binits{A.}},
\bauthor{\bsnm{Lee}, \binits{K.}},
\bauthor{\bsnm{Grover}, \binits{A.}},
\bauthor{\bsnm{Laskin}, \binits{M.}},
\bauthor{\bsnm{Abbeel}, \binits{P.}},
\bauthor{\bsnm{Srinivas}, \binits{A.}},
\bauthor{\bsnm{Mordatch}, \binits{I.}}:
\batitle{Decision transformer: Reinforcement learning via sequence modeling}.
\bjtitle{Advances in neural information processing systems}
\bvolume{34},
\bfpage{15084}--\blpage{15097}
(\byear{2021})
\end{barticle}
\endbibitem

\bibitem[\protect\citeauthoryear{Janner et~al.}{2021}]{janner2021offline}
\begin{barticle}
\bauthor{\bsnm{Janner}, \binits{M.}},
\bauthor{\bsnm{Li}, \binits{Q.}},
\bauthor{\bsnm{Levine}, \binits{S.}}:
\batitle{Offline reinforcement learning as one big sequence modeling problem}.
\bjtitle{Advances in neural information processing systems}
\bvolume{34},
\bfpage{1273}--\blpage{1286}
(\byear{2021})
\end{barticle}
\endbibitem

\bibitem[\protect\citeauthoryear{He et~al.}{2022}]{he2022masked}
\begin{bchapter}
\bauthor{\bsnm{He}, \binits{K.}},
\bauthor{\bsnm{Chen}, \binits{X.}},
\bauthor{\bsnm{Xie}, \binits{S.}},
\bauthor{\bsnm{Li}, \binits{Y.}},
\bauthor{\bsnm{Doll{\'a}r}, \binits{P.}},
\bauthor{\bsnm{Girshick}, \binits{R.}}:
\bctitle{Masked autoencoders are scalable vision learners}.
In: \bbtitle{Proceedings of the IEEE/CVF Conference on Computer Vision and Pattern Recognition},
pp. \bfpage{16000}--\blpage{16009}
(\byear{2022})
\end{bchapter}
\endbibitem

\bibitem[\protect\citeauthoryear{Brohan et~al.}{2023}]{brohan2023can}
\begin{bchapter}
\bauthor{\bsnm{Brohan}, \binits{A.}},
\bauthor{\bsnm{Chebotar}, \binits{Y.}},
\bauthor{\bsnm{Finn}, \binits{C.}},
\bauthor{\bsnm{Hausman}, \binits{K.}},
\bauthor{\bsnm{Herzog}, \binits{A.}},
\bauthor{\bsnm{Ho}, \binits{D.}},
\bauthor{\bsnm{Ibarz}, \binits{J.}},
\bauthor{\bsnm{Irpan}, \binits{A.}},
\bauthor{\bsnm{Jang}, \binits{E.}},
\bauthor{\bsnm{Julian}, \binits{R.}}, \betal:
\bctitle{Do as i can, not as i say: Grounding language in robotic affordances}.
In: \bbtitle{Conference on Robot Learning},
pp. \bfpage{287}--\blpage{318}
(\byear{2023}).
\bcomment{PMLR}
\end{bchapter}
\endbibitem

\bibitem[\protect\citeauthoryear{Brohan et~al.}{2022}]{brohan2022rt}
\begin{botherref}
\oauthor{\bsnm{Brohan}, \binits{A.}},
\oauthor{\bsnm{Brown}, \binits{N.}},
\oauthor{\bsnm{Carbajal}, \binits{J.}},
\oauthor{\bsnm{Chebotar}, \binits{Y.}},
\oauthor{\bsnm{Dabis}, \binits{J.}},
\oauthor{\bsnm{Finn}, \binits{C.}},
\oauthor{\bsnm{Gopalakrishnan}, \binits{K.}},
\oauthor{\bsnm{Hausman}, \binits{K.}},
\oauthor{\bsnm{Herzog}, \binits{A.}},
\oauthor{\bsnm{Hsu}, \binits{J.}}, et al.:
Rt-1: Robotics transformer for real-world control at scale.
arXiv preprint arXiv:2212.06817
(2022)
\end{botherref}
\endbibitem

\bibitem[\protect\citeauthoryear{Ye et~al.}{2023}]{ye2023improved}
\begin{botherref}
\oauthor{\bsnm{Ye}, \binits{Y.}},
\oauthor{\bsnm{You}, \binits{H.}},
\oauthor{\bsnm{Du}, \binits{J.}}:
Improved trust in human-robot collaboration with chatgpt.
IEEE Access
(2023)
\end{botherref}
\endbibitem

\bibitem[\protect\citeauthoryear{Jin et~al.}{2024}]{jin2024robotgpt}
\begin{botherref}
\oauthor{\bsnm{Jin}, \binits{Y.}},
\oauthor{\bsnm{Li}, \binits{D.}},
\oauthor{\bsnm{Yong}, \binits{A.}},
\oauthor{\bsnm{Shi}, \binits{J.}},
\oauthor{\bsnm{Hao}, \binits{P.}},
\oauthor{\bsnm{Sun}, \binits{F.}},
\oauthor{\bsnm{Zhang}, \binits{J.}},
\oauthor{\bsnm{Fang}, \binits{B.}}:
Robotgpt: Robot manipulation learning from chatgpt.
IEEE Robotics and Automation Letters
(2024)
\end{botherref}
\endbibitem

\bibitem[\protect\citeauthoryear{Brockman et~al.}{2016}]{brockman2016openai}
\begin{botherref}
\oauthor{\bsnm{Brockman}, \binits{G.}},
\oauthor{\bsnm{Cheung}, \binits{V.}},
\oauthor{\bsnm{Pettersson}, \binits{L.}},
\oauthor{\bsnm{Schneider}, \binits{J.}},
\oauthor{\bsnm{Schulman}, \binits{J.}},
\oauthor{\bsnm{Tang}, \binits{J.}},
\oauthor{\bsnm{Zaremba}, \binits{W.}}:
Openai gym.
arXiv preprint arXiv:1606.01540
(2016)
\end{botherref}
\endbibitem

\bibitem[\protect\citeauthoryear{Quigley et~al.}{2009}]{quigley2009ros}
\begin{bchapter}
\bauthor{\bsnm{Quigley}, \binits{M.}},
\bauthor{\bsnm{Conley}, \binits{K.}},
\bauthor{\bsnm{Gerkey}, \binits{B.}},
\bauthor{\bsnm{Faust}, \binits{J.}},
\bauthor{\bsnm{Foote}, \binits{T.}},
\bauthor{\bsnm{Leibs}, \binits{J.}},
\bauthor{\bsnm{Wheeler}, \binits{R.}},
\bauthor{\bsnm{Ng}, \binits{A.Y.}}, \betal:
\bctitle{Ros: an open-source robot operating system}.
In: \bbtitle{ICRA Workshop on Open Source Software},
vol. \bseriesno{3},
p. \bfpage{5}
(\byear{2009}).
\bcomment{Kobe, Japan}
\end{bchapter}
\endbibitem

\bibitem[\protect\citeauthoryear{Koenig and Howard}{2004}]{koenig2004design}
\begin{bchapter}
\bauthor{\bsnm{Koenig}, \binits{N.}},
\bauthor{\bsnm{Howard}, \binits{A.}}:
\bctitle{Design and use paradigms for gazebo, an open-source multi-robot simulator}.
In: \bbtitle{2004 IEEE/RSJ International Conference on Intelligent Robots and Systems (IROS)(IEEE Cat. No. 04CH37566)},
vol. \bseriesno{3},
pp. \bfpage{2149}--\blpage{2154}
(\byear{2004}).
\bcomment{IEEE}
\end{bchapter}
\endbibitem

\bibitem[\protect\citeauthoryear{Schulman et~al.}{2017}]{schulman2017proximal}
\begin{botherref}
\oauthor{\bsnm{Schulman}, \binits{J.}},
\oauthor{\bsnm{Wolski}, \binits{F.}},
\oauthor{\bsnm{Dhariwal}, \binits{P.}},
\oauthor{\bsnm{Radford}, \binits{A.}},
\oauthor{\bsnm{Klimov}, \binits{O.}}:
Proximal policy optimization algorithms.
arXiv preprint arXiv:1707.06347
(2017)
\end{botherref}
\endbibitem

\bibitem[\protect\citeauthoryear{Schulman et~al.}{2015}]{schulman2015trust}
\begin{bchapter}
\bauthor{\bsnm{Schulman}, \binits{J.}},
\bauthor{\bsnm{Levine}, \binits{S.}},
\bauthor{\bsnm{Abbeel}, \binits{P.}},
\bauthor{\bsnm{Jordan}, \binits{M.}},
\bauthor{\bsnm{Moritz}, \binits{P.}}:
\bctitle{Trust region policy optimization}.
In: \bbtitle{International Conference on Machine Learning},
pp. \bfpage{1889}--\blpage{1897}
(\byear{2015}).
\bcomment{PMLR}
\end{bchapter}
\endbibitem

\bibitem[\protect\citeauthoryear{Yuan et~al.}{2022}]{yuan2022general}
\begin{bchapter}
\bauthor{\bsnm{Yuan}, \binits{R.}},
\bauthor{\bsnm{Gower}, \binits{R.M.}},
\bauthor{\bsnm{Lazaric}, \binits{A.}}:
\bctitle{A general sample complexity analysis of vanilla policy gradient}.
In: \bbtitle{International Conference on Artificial Intelligence and Statistics},
pp. \bfpage{3332}--\blpage{3380}
(\byear{2022}).
\bcomment{PMLR}
\end{bchapter}
\endbibitem

\bibitem[\protect\citeauthoryear{Lillicrap et~al.}{2015}]{ddpg}
\begin{botherref}
\oauthor{\bsnm{Lillicrap}, \binits{T.P.}},
\oauthor{\bsnm{Hunt}, \binits{J.J.}},
\oauthor{\bsnm{Pritzel}, \binits{A.}},
\oauthor{\bsnm{Heess}, \binits{N.}},
\oauthor{\bsnm{Erez}, \binits{T.}},
\oauthor{\bsnm{Tassa}, \binits{Y.}},
\oauthor{\bsnm{Silver}, \binits{D.}},
\oauthor{\bsnm{Wierstra}, \binits{D.}}:
Continuous control with deep reinforcement learning.
arXiv preprint arXiv:1509.02971
(2015)
\end{botherref}
\endbibitem

\bibitem[\protect\citeauthoryear{Dankwa and Zheng}{2019}]{dankwa2019twin}
\begin{bchapter}
\bauthor{\bsnm{Dankwa}, \binits{S.}},
\bauthor{\bsnm{Zheng}, \binits{W.}}:
\bctitle{Twin-delayed ddpg: A deep reinforcement learning technique to model a continuous movement of an intelligent robot agent}.
In: \bbtitle{Proceedings of the 3rd International Conference on Vision, Image and Signal Processing},
pp. \bfpage{1}--\blpage{5}
(\byear{2019})
\end{bchapter}
\endbibitem

\bibitem[\protect\citeauthoryear{}{}]{ar_track_alvar}
\begin{botherref}
{A ROS wrapper for Alvar, an open source AR tag tracking library}.
\url{https://github.com/ros-perception/ar_track_alvar}
\end{botherref}
\endbibitem

\bibitem[\protect\citeauthoryear{Garrido-Jurado et~al.}{2014}]{garrido2014automatic}
\begin{barticle}
\bauthor{\bsnm{Garrido-Jurado}, \binits{S.}},
\bauthor{\bsnm{Mu{\~n}oz-Salinas}, \binits{R.}},
\bauthor{\bsnm{Madrid-Cuevas}, \binits{F.J.}},
\bauthor{\bsnm{Mar{\'\i}n-Jim{\'e}nez}, \binits{M.J.}}:
\batitle{Automatic generation and detection of highly reliable fiducial markers under occlusion}.
\bjtitle{Pattern Recognition}
\bvolume{47}(\bissue{6}),
\bfpage{2280}--\blpage{2292}
(\byear{2014})
\end{barticle}
\endbibitem

\bibitem[\protect\citeauthoryear{Sucan et~al.}{2012}]{sucan2012open}
\begin{barticle}
\bauthor{\bsnm{Sucan}, \binits{I.A.}},
\bauthor{\bsnm{Moll}, \binits{M.}},
\bauthor{\bsnm{Kavraki}, \binits{L.E.}}:
\batitle{The open motion planning library}.
\bjtitle{IEEE Robotics \& Automation Magazine}
\bvolume{19}(\bissue{4}),
\bfpage{72}--\blpage{82}
(\byear{2012})
\end{barticle}
\endbibitem

\bibitem[\protect\citeauthoryear{Achiam et~al.}{2023}]{achiam2023gpt}
\begin{botherref}
\oauthor{\bsnm{Achiam}, \binits{J.}},
\oauthor{\bsnm{Adler}, \binits{S.}},
\oauthor{\bsnm{Agarwal}, \binits{S.}},
\oauthor{\bsnm{Ahmad}, \binits{L.}},
\oauthor{\bsnm{Akkaya}, \binits{I.}},
\oauthor{\bsnm{Aleman}, \binits{F.L.}},
\oauthor{\bsnm{Almeida}, \binits{D.}},
\oauthor{\bsnm{Altenschmidt}, \binits{J.}},
\oauthor{\bsnm{Altman}, \binits{S.}},
\oauthor{\bsnm{Anadkat}, \binits{S.}}, et al.:
Gpt-4 technical report.
arXiv preprint arXiv:2303.08774
(2023)
\end{botherref}
\endbibitem

\bibitem[\protect\citeauthoryear{Takaya et~al.}{2016}]{Rviz}
\begin{bchapter}
\bauthor{\bsnm{Takaya}, \binits{K.}},
\bauthor{\bsnm{Asai}, \binits{T.}},
\bauthor{\bsnm{Kroumov}, \binits{V.}},
\bauthor{\bsnm{Smarandache}, \binits{F.}}:
\bctitle{Simulation environment for mobile robots testing using ros and gazebo}.
In: \bbtitle{2016 20th International Conference on System Theory, Control and Computing (ICSTCC)},
pp. \bfpage{96}--\blpage{101}
(\byear{2016}).
\bcomment{IEEE}
\end{bchapter}
\endbibitem

\bibitem[\protect\citeauthoryear{Aulinas et~al.}{2008}]{aulinas2008slam}
\begin{botherref}
\oauthor{\bsnm{Aulinas}, \binits{J.}},
\oauthor{\bsnm{Petillot}, \binits{Y.}},
\oauthor{\bsnm{Salvi}, \binits{J.}},
\oauthor{\bsnm{Llad{\'o}}, \binits{X.}}:
The slam problem: a survey.
Artificial Intelligence Research and Development,
363--371
(2008)
\end{botherref}
\endbibitem

\bibitem[\protect\citeauthoryear{}{}]{rqt}
\begin{botherref}
{Qt-based framework for GUI development for ROS}.
\url{http://wiki.ros.org/rqt}
\end{botherref}
\endbibitem

\end{thebibliography}

\end{document}